\documentclass[10pt,twocolumn,letterpaper]{article}

\usepackage{iccv}
\usepackage{times}
\usepackage{epsfig}
\usepackage{graphicx}
\usepackage{amsmath, bm}
\usepackage{amssymb}
\usepackage{booktabs}
\usepackage{multirow}
\usepackage{pifont}
\usepackage[table,xcdraw]{xcolor}
\usepackage{enumitem}
\usepackage[accsupp]{axessibility} % Improves PDF readability for those with disabilities

\usepackage{xr}
\makeatletter

\newcommand*{\addFileDependency}[1]{% argument=file name and extension
\typeout{(#1)}% latexmk will find this if $recorder=0
\@addtofilelist{#1}
\IfFileExists{#1}{}{\typeout{No file #1.}}
}\makeatother

\newcommand*{\myexternaldocument}[1]{%
\externaldocument{#1}%
\addFileDependency{#1.tex}%
\addFileDependency{#1.aux}%
}

\myexternaldocument{egpaper_supp}
\usepackage[pagebackref=true,breaklinks=true,letterpaper=true,colorlinks,bookmarks=false]{hyperref}
\RequirePackage[format=plain,labelformat=simple,labelsep=period,font=small,compatibility=false]{caption}
\RequirePackage[font=footnotesize,skip=3pt,subrefformat=parens]{subcaption}

\usepackage[capitalize]{cleveref}
\crefname{section}{Sec.}{Secs.}
\Crefname{section}{Section}{Sections}
\Crefname{table}{Table}{Tables}
\crefname{table}{Tab.}{Tabs.}

\usepackage[breaklinks=true,bookmarks=false]{hyperref}

\iccvfinalcopy % *** Uncomment this line for the final submission

\def\iccvPaperID{****} % *** Enter the ICCV Paper ID here

\ificcvfinal\fi

\begin{document}

%%%%%%%%% TITLE
\title{Dynamic Residual Classifier for Class Incremental Learning}

\author{Xiuwei Chen\textsuperscript{1}, Xiaobin Chang\textsuperscript{1,2,3}\thanks{indicates corresponding author.}\\
% School of Artificial Intelligence, Sun Yat-sen University\\
% China\\
% {\tt\small chenxw83@mail2.sysu.edu.cn}
% \and
% Xiaobin Chang\\
$^1$School of Artificial Intelligence, Sun Yat-sen University, China\\
$^2$Guangdong Key Laboratory of Big Data Analysis and Processing, Guangzhou 510006, P.R.China\\
$^3$Key Laboratory of Machine Intelligence and Advanced Computing, Ministry of Education, China\\
{\tt\small chenxw83@mail2.sysu.edu.cn, changxb3@mail.sysu.edu.cn}
}

\maketitle
\ificcvfinal\thispagestyle{empty}\fi

%%%%%%%%% ABSTRACT
\begin{abstract}

The rehearsal strategy is widely used to alleviate the catastrophic forgetting problem in class incremental learning (CIL) by preserving limited exemplars from previous tasks.
With imbalanced sample numbers between old and new classes, the classifier learning can be biased.
% With the imbalanced sample numbers between old and new classes, the learned classifiers can be biased.
% data samples
% biased classifiers can be learned and harm the performance.
Existing CIL methods exploit the long-tailed (LT) recognition techniques, e.g., the adjusted losses and the data re-sampling methods, to handle the data imbalance issue within each increment task.
% independently.
% handle the data imbalance issue within each increment task with either the adjusted losses or the re-sampling methods of long-trailed (LT) recognition.
% Existing CIL methods handle the data imbalance issue within each increment task with either the adjusted losses or the re-sampling methods of long-trailed (LT) recognition.
% Existing CIL methods separately handle the data imbalance within each increment task with either the adjusted losses or the re-sampling methods of long-trailed (LT) recognition.
% Existing CIL methods adopt the adjusted losses and data re-sampling methods of long-trailed (LT) recognition to handle the data imbalance within each task.
% To handle such data imbalance issues of CIL, existing methods adopt the adjusted losses and/or data re-samplings from Long-trailed (LT) recognition. % within each task.
In this work, the dynamic nature of data imbalance in CIL is shown and a novel Dynamic Residual Classifier (DRC) is proposed to handle this challenging scenario.
% In this work, we show the dynamic nature of data imbalance in CIL and propose a novel Dynamic Residual Classifier (DRC) to handle this challenging scenario.
% A novel Dynamic Residual Classifier (DRC) is proposed to handle this challenging scenario.
% To handle this challenging scenario, the novel Dynamic Residual Classifier (DRC) based on a residual fusion mechanism and task-specific branch layers is proposed.
% Based on a residual fusion mechanism, the novel Dynamic Residual Classifier (DRC) is then proposed for more balanced learning across tasks.
% A novel Dynamic Residual Classifier (DRC) based on a residual fusion mechanism is then proposed for more balanced learning across tasks.
Specifically, DRC is built upon a recent advance residual classifier with the branch layer merging to handle the model-growing problem.
% and
% the classifier of the new task is continually learned with the previously learned classifiers fixed.
% a simple yet effective branch layer merging process is designed to handle the model-growing problem.
% in CIL.
% Moreover, DRC can be generalizable to different pipelines and settings. 
% Moreover, DRC is compatible with different distillation pipelines and brings substantial improvements.
Moreover, DRC is compatible with different CIL pipelines and substantially improves them.
Combining DRC with the model adaptation and fusion (MAF) pipeline, this method achieves state-of-the-art results on both the conventional CIL and the LT-CIL benchmarks.
% Combining DRC with the model adaptation and fusion (MAF) pipeline, this method achieves state-of-the-art results on both the conventional CIL and the long-tailed CIL (LT-CIL) benchmarks.
% Combining DRC with the model adaptation and fusion (MAF) pipeline, this method achieves state-of-the-art results under both the conventional CIL and the long-tailed CIL (LT-CIL) settings.
% Combining DRC with a backbone pipeline, the resulting model can achieve state-of-the-art results under both the conventional CIL and long-tailed CIL (LT-CIL) settings.
% Combining DRC with the model adaptation and fusion (MAF) pipeline, the resulting model, MAFDRC, achieves state-of-the-art results under both the conventional CIL and long-tailed CIL (LT-CIL) settings.
Extensive experiments are also conducted for a detailed analysis. The code is publicly available\footnote{
\url{https://github.com/chen-xw/DRC-CIL}
}.

\end{abstract}

%%%%%%%%% BODY TEXT
\section{Introduction}
\label{sec:intro}

%% 1. catastrophic forgetting problem of DNN in CIL
Deep models are prone to forgetting previously learned knowledge when sequentially fine-tuned on different tasks.
% with disjoint classes.
Severe performance degradation on the old tasks can be observed.
It is also known as catastrophic forgetting~\cite{catastrophic,french2002using}.
Class incremental learning (CIL) methods~\cite{continual,masana2020class} aim to handle this issue and equip deep models with the capacity to continuously learn new categories without forgetting the old ones.
The rehearsal strategy~\cite{icarl,wa,rolnick2019experience,foster,der} has been widely used to achieve this goal. Specifically, a limited amount of exemplars from previous tasks are stored in a memory buffer and replayed when learning new tasks.

% AI agents, e.g., deep neural networks (DNNs), deployed in the real world face an ever-changing environment, i.e., with new concepts and categories continually emerging~\cite{shaheen2022continual, geiger2012we, li2022coda}.
% Therefore, class incremental learning (CIL)~\cite{continual, masana2020class, icarl} has attracted much attention as it aims to equip deep models with the capacity to continuously handle new categories.
% However, learning to discriminate sets of disjoint classes sequentially with deep models is challenging, as they are prone to entirely forgetting the previous knowledge, thus resulting in severe performance degradation of old tasks. It is known as catastrophic forgetting~\cite{catastrophic}.

\begin{figure}[t]
  \centering
%   \fbox{\rule{0pt}{2in} \rule{0.9\linewidth}{0pt}}
  \includegraphics[width=1.0\linewidth]{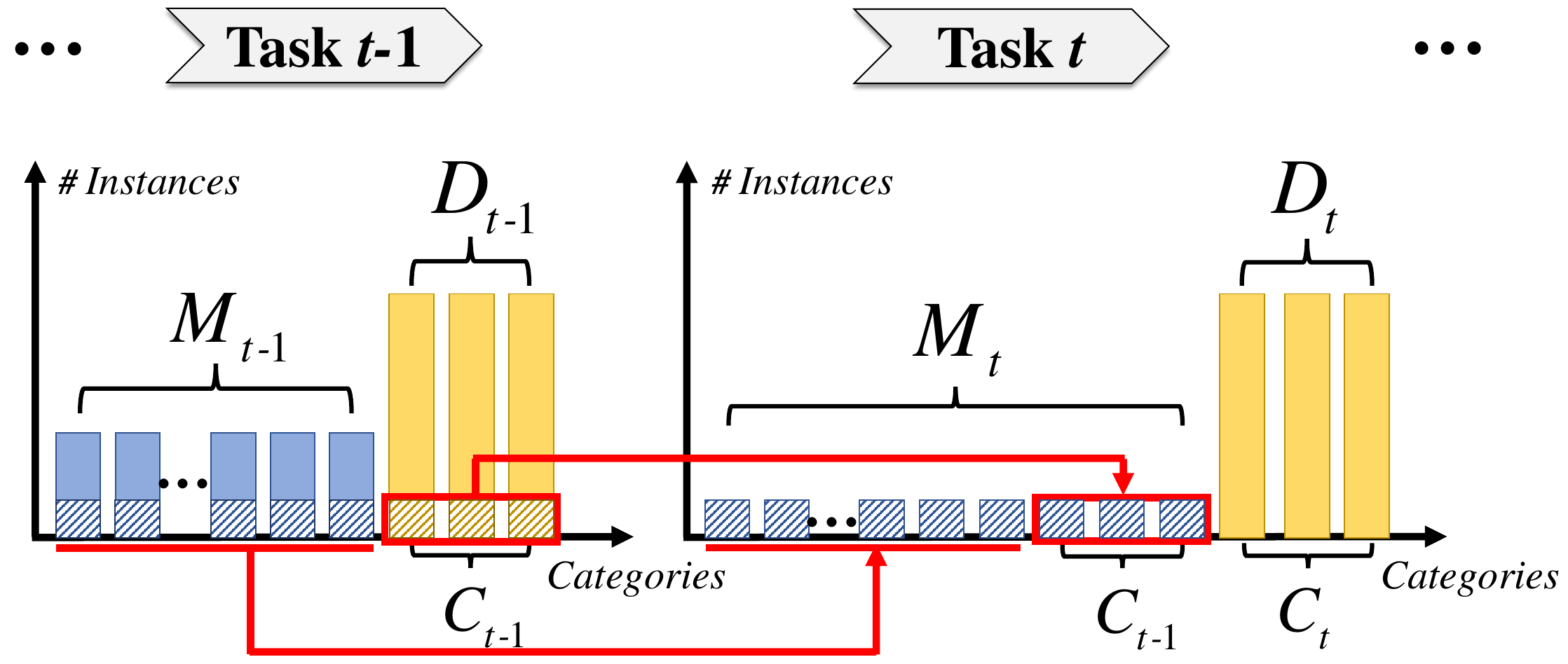}
   \caption{
   Data imbalance of CIL. With the exemplars of previous tasks buffered, the training data within each task is imbalanced.
   As the task increment proceeds, more categories appear in a fixed-size memory. Such imbalance becomes more severe.
   % Dynamic data imbalance in CIL. As the task increment proceeds, from $t-1$ to $t$, more categories appear in a fixed-size memory $M_t$. The imbalance is thus more severe.
   }
   \label{fig:data_imba}
\end{figure}

% To alleviate the catatrophic forgetting, the rehearsal strategy\cite{icarl} is exploited. With only limited samples from previous tasks stored in a relatively small memory buffer and used for the new task. 
% \cite{icarl} saves an extra exemplar set that can  best approximate class means in the learned feature space, and includes it into the model updating. \cite{chaudhry2018riemannian,bang2021rainbow,aljundi2019gradient} propose corresponding sampling measures to select the informative ones.

Due to the relatively small size of the memory buffer, the training samples of a new class are far more than the old ones.
% , as depicted in~\cref{fig:data_imba}.
Therefore, adopting the rehearsal strategy can introduce the data imbalance problem to CIL.
Two kinds of long-tailed recognition techniques, the adjusted losses~\cite{menon2020long,cui2019class} and the data re-sampling~\cite{kang2019decoupling}, are exploited by many CIL methods~\cite{hou2019learning,foster,castro2018end,der} to learn the classifier  with less bias.
These methods alleviate the data imbalance within each increment task independently.
% These methods mainly alleviate the static data imbalance within each increment task.
% Such long-tailed methods independently handle the data imbalance within each task.
% to learn more balanced classifiers.
% a more balanced classifier for each task.
% handle the imbalance issue within each task.
% within each task.

However, the data imbalance in CIL is dynamic and becomes more extreme as the task increment proceeds, as illustrated in~\cref{fig:data_imba}.
A novel dynamic residual classifier (DRC) is proposed in this work to handle this challenging scenario.
Inspired by the recent advance residual classifier (RC)~\cite{cui2022reslt}, a lightweight branch layer is inserted before the classifier to encode the task-specific knowledge.
This new architecture enables the residual fusion of classifier outputs to alleviate the data imbalance effectively.
% This new architecture enables the residual fusion of classifier outputs to alleviate the data imbalance issue effectively.
% and is complementary to previous efforts.
However, directly applying RC for CIL leads to the model-growing problem, i.e., the growing overhead from the additional branch layers assigned to the new tasks.
% However, directly applying RC for CIL leads to the model-growing problem, i.e., the growing overhead of model parameters from additional branch layers assigned to the new tasks.
% The proposed DRC handles this dynamic increment issue with branch layer merging.
The proposed DRC handles this dynamic increment issue via the simple yet effective branch layer merging.

\begin{figure}[t]
  \centering
%   \fbox{\rule{0pt}{2in} \rule{0.9\linewidth}{0pt}}
  \includegraphics[width=1.0\linewidth]{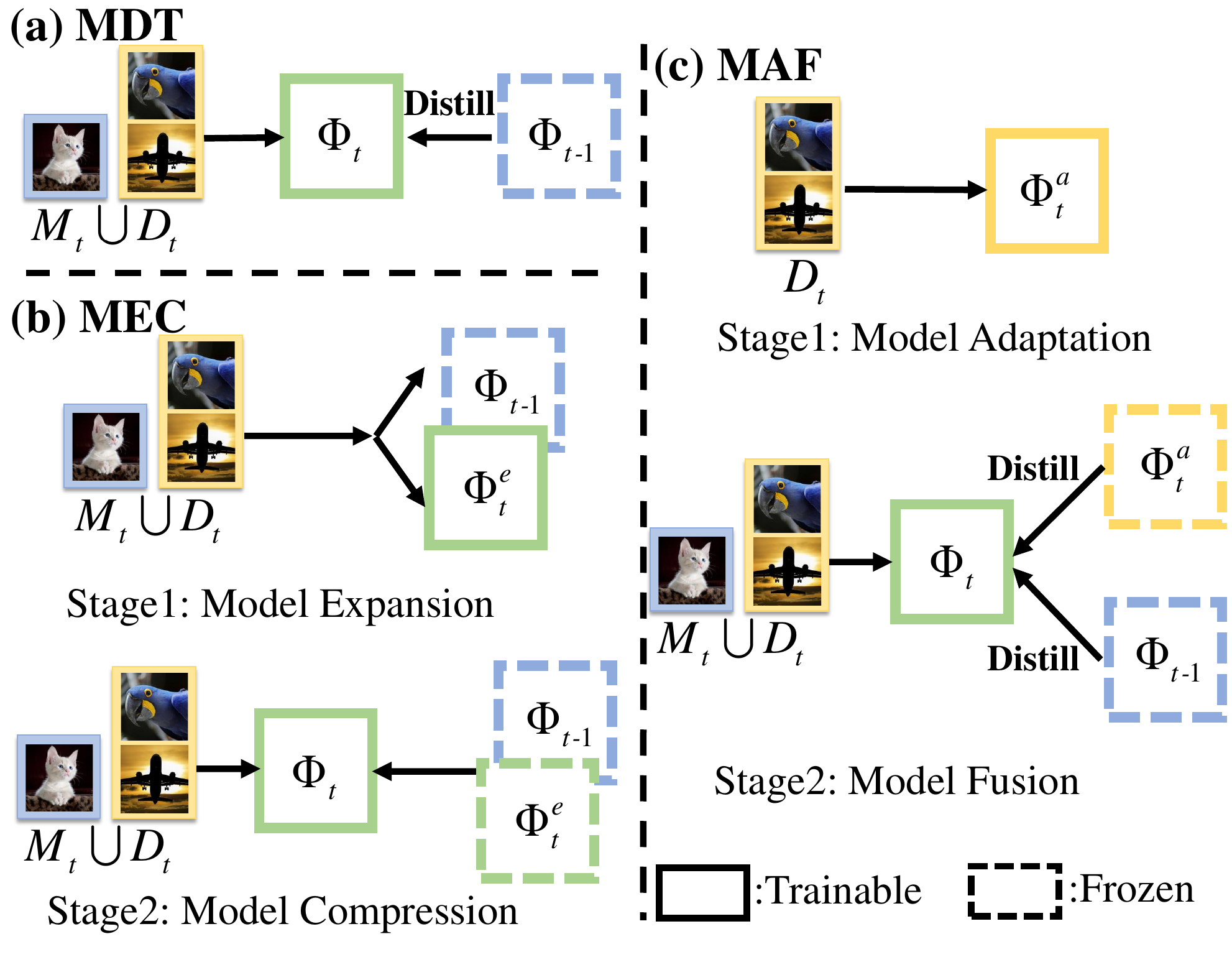}
   \caption{
   % Illustrations of three distillation pipelines of CIL.
   Illustrations of three CIL pipelines.
   (a) is the Model Direct Transfer (MDT). (b) depicts the Model Expansion and Compression (MEC). (c) shows the Model Adaptation and Fusion (MAF).
   % Rehearsal is used with new data $D_t$ and memory data $M_t$ of task $t$.
   $\Phi_t$ indicates the classification model of task $t$.
   % The classification model $\Phi_t$ is trained with $D_t$ and $M_t$ at task $t$.
   $D_t$ and $M_t$ represent the new data and memory data of task $t$ respectively.
   Blue indicates old knowledge, yellow indicates new one and green represents a mixture of both.}
   \label{fig:frame}
\end{figure}

DRC is directly applicable to different CIL pipelines by simply replacing the fully connected (fc) classifiers.
% DRC is applicable to different CIL pipelines by simply replacing the fc classifier head with it.
Three typical CIL pipelines, i.e., the Model Direct Transfer (MDT)~\cite{icarl,podnet}, the Model Expansion and Compression (MEC)~\cite{foster,der} and the Model Adaptation and Fusion (MAF)~\cite{hou2018lifelong,choi2021dual}, as shown in \cref{fig:frame}, are chosen to be combined with DRC.
They can consistently benefit from such combinations, with clear improvements observed.
% Three typical CIL pipelines, i.e., the Model Direct Transfer (MDT)~\cite{icarl,podnet}, the Model Expansion and Compression (MEC)~\cite{foster} and the Model Adaptation and Fusion (MAF)~\cite{choi2021dual,hou2018lifelong}, as depicted in \cref{fig:frame}, are chosen for validation.
The details and comparisons on such combinations are given in \cref{sec:mafdrc}.
% The details and comparisons about their combinations are in \cref{sec:mafdrc}.
% Combining with DRC can bring them substantial improvements.
More importantly, DRC is most compatible with MAF among the three pipelines. The resulting MAFDRC method achieves state-of-the-art performance under both conventional CIL and long-tailed CIL (LT-CIL) settings.
% , as detailed and compared in \cref{sec:mafdrc}.
% The resulting method, denoted as MAFDRC, achieves state-of-the-art performance under both the conventional CIL and the long-tailed CIL (LT-CIL) settings.
Extensive analyzes are conducted to provide insights into each part.
%% Conclusion
% The main contributions of this paper are as follows.
% The main contributions of this paper are summarized as follows.
The main contributions
% of this paper 
are three-fold:
% \begin{itemize}
\begin{itemize}[noitemsep]
  \item We show the data imbalance issue in CIL rehearsal is dynamic across tasks rather than static within each task. The proposed dynamic residual classifier (DRC) aims to handle this challenging scenario from the perspective of classifier architecture, which is complementary to existing efforts in CIL;
  \item The branch layer architecture and residual fusion mechanism from a recent long-tailed classifier (RC)~\cite{cui2022reslt} are adopted by DRC to alleviate the negative impact of data imbalance on CIL for the first time.
  More importantly, the model-growing problem of the vanilla RC under the CIL setting is handled with the simple yet effective branch layer merging in DRC;
  \item The proposed DRC is generalizable.
  On the one hand, incorporating DRC brings clear improvements to different CIL pipelines.
  On the other hand, the effectiveness of DRC is demonstrated in both the CIL and the LT-CIL settings.
\end{itemize}

% The main contributions of this paper are two-fold.
% (1) It is the first attempt to handle the data imbalance in CIL with the classifier architectures. In particular, a novel classifier DRC is proposed with the CIL dynamic procedure considered.
% (2) The novel classifier DRC is very generalizations not only for different pipelines but on a new experiment setting long-tailed CIL scenerios.
% The effectiveness of MAFDRC is demonstrated by the state-of-the-art performance on three large-scale CIL benchmarks. Extensive studies are also conducted to provide insights into each part.
% (1) The proposed model adaptation and fusion (MAF) is a more consistent and integrated CIL pipeline than its counterparts.
% The model learned with MAF thus achieves a better plasticity-stability trade-off and results in superior performance.
% (2) It is the first attempt to handle the data imbalance in CIL with the classifier architectures. In particular, a novel classifier DRC is proposed with the CIL dynamic procedure considered.
% The effectiveness of MAFDRC is demonstrated by the state-of-the-art performance on three large-scale CIL benchmarks. Extensive studies are also conducted to provide insights into each part.

%------------------------------------------------------------------------
\section{Related Work}
\label{sec:relat}

\noindent {\bf Class Incremental Learning} (CIL) is one of the major settings in continual learning~\cite{van2019three,de2021continual}. It aims to equip deep models with the capacity to continually learn from a sequence of tasks with disjoint classes and avoids the Catastrophic Forgetting~\cite{catastrophic} of previously learned knowledge.
% Three main approaches have been proposed to achieve this purpose.
{\bf Rehearsal-based methods} ~\cite{icarl,gem,wa,wu2019large,rolnick2019experience,isele2018selective} preserve very limited exemplars from previous tasks and replaying them in the new task to defy forgetting.
Exemplars are selected by different strategies.
iCaRL~\cite{icarl} stores a subset of samples per class by selecting the good approximations to class means in the feature space.
RWalk~\cite{chaudhry2018riemannian} selects exemplars with higher entropy or those close to the classification boundary. Instead of storing raw data samples, DGR~\cite{generative} exploits the synthetic instances from a generative model~\cite{goodfellow2020generative}.
In this work, we use the rehearsal strategy as in~\cite{icarl}. %iCaRL
% On the contrary, no exemplar from the previous task can be exploited by the Rehearsal-free methods~\cite{zhu2022self, toldo2022bring, petit2023fetril}.
% ~\cite{zhu2021prototype, zhu2022self, toldo2022bring, zhu2021class, petit2023fetril}.
% for CIL.
% are also widely studied for CIL.
% Classifier Learning for CIL
The data imbalance caused by limited rehearsal memory is a challenge for \textbf{Classifier Learning in CIL}.
% The data imbalance caused by limited rehearsal memory is the main challenge for \textbf{Classifier Learning in CIL}.
To learn a less biased classifier, the adjusted losses~\cite{menon2020long,cui2019class} are exploited by existing CIL methods.
A margin ranking loss proposed by \cite{hou2019learning} encourages the old and new classes to be better separated and avoid ambiguities.
The adjusted classification loss of FOSTER~\cite{foster} aims to re-balance the logits of the rare (old) and dominant (new) classes.
Based on a balanced training subset sampled \cite{kang2019decoupling}, an independent classifier learning stage is introduced to alleviate the impact of data imbalance.
% Based on a balanced training subset sampled, an independent classifier learning stage is introduced after representation learning to avoid the impact of data imbalance \cite{kang2019decoupling}.
For example, EEIL~\cite{castro2018end} finetunes its classifier while DER~\cite{der} trains a new one from scratch with such balanced data.
% A balanced training subset can be sampled and used only for learning the classifier to avoid the impact of data imbalance \cite{kang2019decoupling}.
% The learning of a classifier can be carried out on a balanced training subset and decoupled with the representation learning.
% In addition, the data imbalance issue can be handled by decoupling the classifier learning from the representation learning \cite{kang2019decoupling}. Based on a balanced training subset sampled, the classifier is finetuned in EEIL~\cite{castro2018end} while a new one is trained from scratch in DER~\cite{der}.
% EEIL~\cite{castro2018end} finetunes its classifier while DER~\cite{der} trains a new one from scratch.
Moreover, different post-hoc corrections are applied to the classifiers learned from the imbalanced data.
The output logits of the new classifier are rescaled by a simple affine function in BiC~\cite{wu2019large}.
The norms of the classifier weight vectors for the new and old classes are aligned in WA~\cite{wa}.
The proposed dynamic residual classifier (DRC) aims to handle the data imbalance of CIL with a new classifier of dynamic architecture.
% the novel classifier architecture~\cite{cui2022reslt}.
Therefore, it is complementary to the relevant CIL methods mentioned above.
% contribution
% Besides the rehearsal strategy, distillation-based methods~\cite{icarl,wa,podnet,dhar2019learning,choi2021dual,hou2018lifelong} transfer the discriminative knowledge of preceding categories from the old models to the new ones.
% % Besides the rehearsal strategy, distillation-based methods~\cite{icarl,wa,podnet,dhar2019learning,choi2021dual,hou2018lifelong} are used to transfer the discriminative knowledge of preceding categories from the learned old models to the new task ones.
% We focus on three typical CIL pipelines with model distillations, i.e., Model Direct Transfer (MDT)~\cite{icarl,podnet}, Model Expansion and Compression (MEC)~\cite{foster} and Model Adaptation and Fusion (MAF)~\cite{choi2021dual,hou2018lifelong}, as shown in \cref{fig:frame}.

Besides the rehearsal strategy, distillation techniques~\cite{hinton2015distilling,zhao2022decoupled} are used by different CIL pipelines to transfer the discriminative knowledge of preceding categories from the old models to the new ones, resulting in the {\bf Distillation-based methods}~\cite{icarl,podnet,foster,hou2018lifelong,choi2021dual}.
% MDT
Under the Model Direct Transfer (MDT) pipeline, the model is finetuned with both the new data and the distilled knowledge from the retained old model. 
% distillation is used to directly transfer the knowledge from the retained old model to the trainable new model 
The changes in attention maps between the new and old models are penalized via distillation in LwM~\cite{dhar2019learning}.
Distillation also can be conducted on the prediction scores~\cite{icarl} or spatial features~\cite{podnet}.
% Many methods directly use distillation loss to transfer the knowledge of the previous task. 
% For example, iCaRL~\cite{icarl} retains the old model and distills the knowledge from the old model into the new one. 
% LwM~\cite{dhar2019learning} applies the attention distillation loss to limit the changes.
% PODNet~\cite{podnet} uses a spatial-based distillation loss to constrain the evolution of the representation. 
%-----
% MEC (some ME is also the {\bf Parameter Isolation-based methods})
The Model Expansion and Compression (MEC) pipeline consists of two stages.
In the first stage, the old model is retained and expanded with new modules for the learning of a new task. Such a model expansion stage is also known as the parameter isolation methods~\cite{learn,compacting,der,foster,dytox}.
% \cite{learn,pathnet,compacting,2019continual}.
% old frozen model is first expanded with new modules to learn a new task.
% ~\cite{learn,pathnet,compacting,2019continual,serra2018overcoming}
The feature representations from the old frozen model and the newly added one are concatenated and trained on the new task, as in DER~\cite{der} and FOSTER~\cite{foster}.
A dynamic model expansion strategy based on the ViT architecture~\cite{dosovitskiy2020image} is proposed in DyTox~\cite{dytox}.
In the next stage, the model compression, e.g., distillation~\cite{hinton2015distilling} or network pruning~\cite{hat}, is applied to control the size of the expanded model.
% Model compression is then applied in the next stage to control the size of the expanded model. It can be achieved by either model distillation~\cite{hinton2015distilling} or network pruning~\cite{hat}.
% However, only use a model to keep the balance between the contracting goals of incorporating model classes and keeping feature representation unchanged may be difficult. So some methods~\cite{learn,pathnet,compacting,2019continual,serra2018overcoming} dynamically expand new modules to learn new tasks.
% % expand a new model to learn new knowledge while concatenating the previously learned model. % while distilling the knowledge from previous model to the new expanded model. 
% These methods are designed for task incremental learning~\cite{van2019three} and require task-id to activate the corresponding sub-network.
% Recently, a few methods~\cite{der,dytox,foster} have been proposed for CIL and achieved state-of-the-art performance.
% A novel dynamic model expansion strategy is proposed in DyTox~\cite{dytox} based on the ViT architecture~\cite{dosovitskiy2020image}.
% DER~\cite{der} and FOSTER~\cite{foster} concatenate the feature representations to train from the old frozen model and the new added one.
% Model compression~\cite{hat,hinton2015distilling} is then used to control the model size.
%-----
% MAF
Under the Model Adaptation and Fusion (MAF) pipeline, a model optimized on the new task only is obtained at the adaptation stage.
The new knowledge within this adapted model together with the old knowledge from either the exemplars~\cite{hou2018lifelong} or the old model~\cite{choi2021dual} are integrated into a single model via distillation.
% \textcolor{red}{
A neural network is split into two partitions in~\cite{kim2021split} for training the new task separated from the old task and reconnecting them to fuse the knowledge across tasks.
% }
% Some methods also expand a new model but only learn the new knowledge. Then the knowledge of the old and new model is integrated to a single model. 
% Dual-teacher~\cite{choi2021dual} uses dual-teacher information distillation for knowledge distillation from two teachers to one student.
% DR~\cite{hou2018lifelong} distills the knowledge from expert CNN and original CNN to a single model.
%-----
% DRC compatible all three and MAFDRC best 
The proposed DRC is compatible with the three pipelines and clearly improves their performance.
% The DRC can be compatible for all three pipelines and MAF is the most compatible.

\noindent {\bf Long-tailed (LT) Recognition}~\cite{zhang2021deep, yang2022survey} is an active research topic under Data Imbalance~\cite{kaur2019systematic,he2009learning}.
The adjusted losses~\cite{menon2020long,cui2019class,zhang2017range} and the data re-samling~\cite{buda2018systematic,byrd2019effect,devi2017redundancy,kang2019decoupling} are two kinds of important techniques for Long-tailed Recognition.
They are adopted by many CIL methods~\cite{castro2018end,hou2019learning,wa,der,foster} to independently handle the data imbalance within each incremental task, as detailed above.
% However, the data imbalance of CIL can be more challenging than that of LT as the former is dynamic across incremental tasks, as shown in this work.
In this work, we show the data imbalance of CIL can be more challenging than that of LT, as the former is dynamic across incremental tasks while the latter is static.
% In this work, we show the data imbalance of CIL is dynamic rather than static as in LT.
% can be more challenging than LT.
% the dynamic nature of the data imbalance
% By assuming the data imbalance issue of CIL as a sequence of static LT problem  
Recently, a novel classifier architecture~\cite{cui2022reslt} has been proposed for LT recognition.
We find its branch layer architecture and residual fusion mechanism are effective under the CIL setting.
% for handling the data imbalance of CIL.
However, directly applying this classifier for CIL raises the model-growing problem.
The proposed DRC handles it with the branch layer merging and becomes an effective and efficient classifier for CIL.
Moreover, a new CIL setting, long-tailed CIL (LT-CIL), has been proposed in~\cite{liu2022long}, where the new task data obeys a long-tailed distribution as well. Our DRC is found effective in this setting.

\begin{figure*}[t]
  \centering
%   \fbox{\rule{0pt}{2in} \rule{0.9\linewidth}{0pt}}
  % \includegraphics[width=1.0\linewidth]{figs/preliminary.pdf}
  \includegraphics[width=0.9\linewidth]{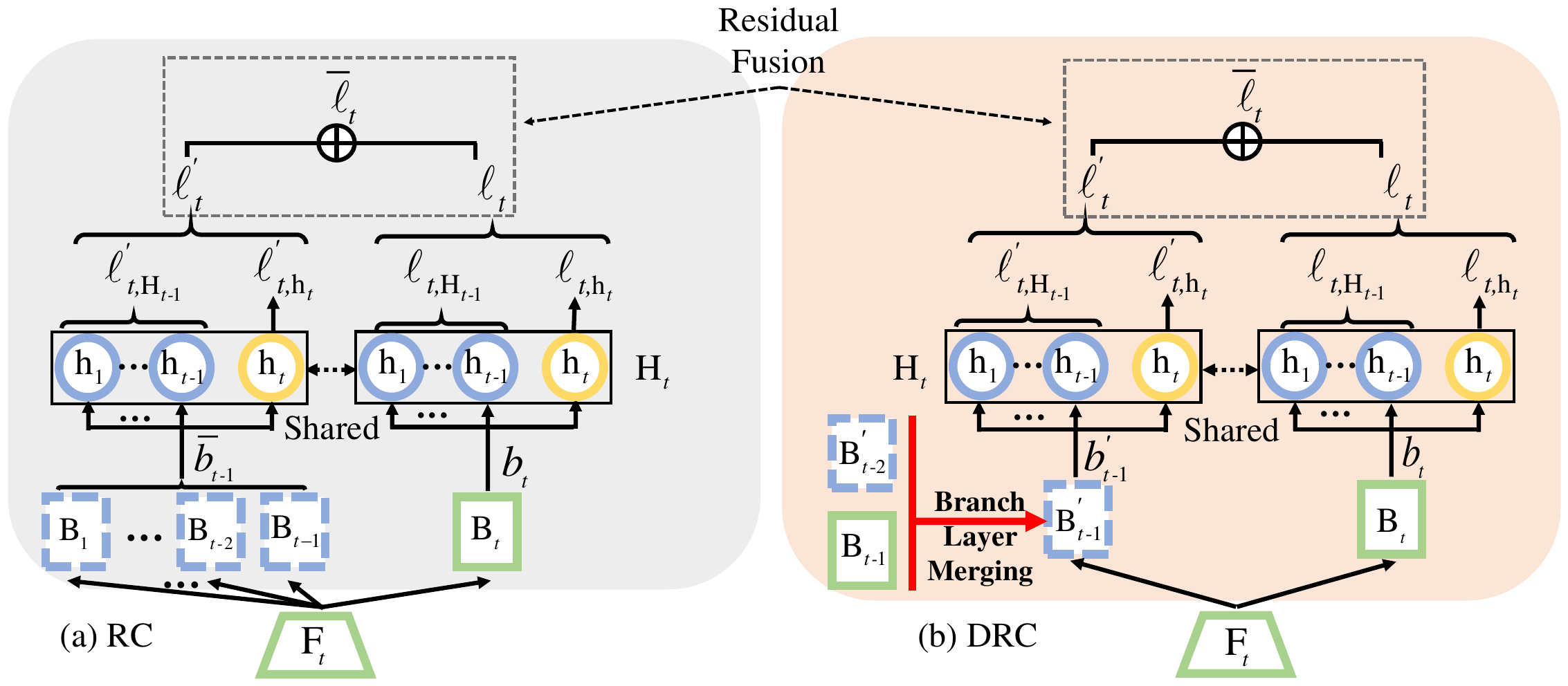}
  \caption{
  Illustrations of the residual classifier (RC) and the proposed dynamic residual classifier (DRC) for CIL at task $t$.
  The residual fusion of overall logits, ${{{\rm{\bar \ell}}}_t}$, is shown.
  $\operatorname{F}_t$ is the backbone feature extractor.
  % The residual classifier and dynamic residual classifier for CIL.
  }
  % \caption{ Outputs of the model.}
  \label{fig:maf parameters}
\end{figure*}

%------------------------------------------------------------------------
% \section{Dynamic Residual Classifier}\label{sec:maf}
% \section{The Proposed Method}\label{sec:maf}
\section{Methodology}\label{sec:maf}

%-------------------------------------------------------------------------
% \subsection{Preliminary}\label{sec:preliminary}

% categories, training data D_t, M_t, M_t features, test set.
In class incremental learning, a model is learned from a sequence of ${T}$ tasks, where the task ${t}$ has a set of $n_t$ different classes ${C_t} = \{ c_{t,1},c_{t,2}, \cdots ,c_{t,n_t}\}$. The classes in different tasks are disjoint, ${C_{i}} \cap {C_j} = \emptyset, \forall i,j \in \{1,...,T\}$.
The training data of task $t$ denotes as ${D_t}$. It contains data tuples in the form of $(x,y)$ where $x$ is the image and $y$ is its ground-truth class label. When training on task ${t}$, the model can only access to ${D_t}$.
With rehearsal applied, data samples from previous tasks are maintained in a memory buffer $M_t$ with a relatively small size, i.e.,
$|{M_t}| \ll |{D_t}|$.
% but fixed 
% $K, K \ll |{D_t}|$. 
$M_t$ is also included in the learning procedure of task $t$.
Updates of the memory buffer are dynamic when the task increment proceeds, as shown in \cref{fig:data_imba}.
During testing, data samples are from all observed classes so far with balanced distributions.

\subsection{Residual Classifier for CIL}\label{sec:rc}
% The classification model $\Phi_t$ is trained with $D_t$ and $M_t$ at task $t$.
The residual classifier (RC) \cite{cui2022reslt} is with the branch layer architecture and residual fusion mechanism. RC can be used to handle the dynamic data imbalance of CIL.
Specifically, the classification model $\Phi_t$ with RC consists of three parts, the feature extractor $\operatorname{F}_t$, the branch layers ${ \operatorname{B}_{1},\cdots,\operatorname{B}_{t} }$ and the classifier heads $\operatorname{H}_{t} = \{ {\operatorname{h}_1},{\operatorname{h}_2}, \cdots ,{\operatorname{h}_{t}} \}$, as illustrated in \cref{fig:maf parameters}(a).
% The feature representation $f \in \mathbb{R}^d$ of input image $x$ is extracted via $\operatorname{F}_t(x)$.
 The feature representation of an input image $x$ is
 \begin{equation}
  f = \operatorname{F}_t(x),
  % \label{eq:}
\end{equation}
where $f \in \mathbb{R}^d$ and $\operatorname{F}_t$ is parameterized with $\theta_t$.
The branch layers are task-specific under the CIL setting. $\operatorname{B}_{1},\cdots,\operatorname{B}_{t-1}$ are inherited from the previous model $\Phi_{t-1}$ to preserve the old knowledge. Therefore, they are frozen at the new task $t$. $\operatorname{B}_{t}$ is for task $t$ and learned with other parts of $\Phi_t$. All branches are lightweight, i.e., the $1\times1$ convolutional layers without bias terms, to alleviate the growing overhead of model parameters. The output of the branch layer $\operatorname{B}_i$ is
\begin{equation}
  b_i = \operatorname{B}_i(f) = \omega_{\operatorname{B}_i} \cdot f,
  % \label{eq:}
\end{equation}
where $\omega_{\operatorname{B}_i}$ is the weight of branch layer and $\cdot$ is matrix multiplication, $i = 1, \cdots, t$. To encode the knowledge of all previous tasks, their outputs are averaged,
\begin{equation}
  \bar{b}_{t-1} = \frac{b_1 + \cdots + b_{t-1}}{t-1},
  % \label{eq:}
\end{equation}
with $t \geq 2$.
$\operatorname{H}_t$ is a set of task-specific classifiers ${ {\operatorname{h}_1},{\operatorname{h}_2}, \cdots ,{\operatorname{h}_{t}} }$, where $\operatorname{h}_i$ corresponds to the classifier of task $i, i=1, \cdots, t$. $\operatorname{h}_i$ is an FC layer with output dimension $|C_i|$.
Different logits of classifiers are computed via
\begin{equation}
% \begin{split}
\left\{ {\begin{array}{*{20}{l}}
\ell^{'}_{t, \operatorname{H}_{t-1}} = \operatorname{H}_{t-1}(\bar{b}_{t-1}),\\
\ell^{'}_{t, h_t} = h_t(\bar{b}_{t-1}),\\
\ell_{t, \operatorname{H}_{t-1}} = \operatorname{H}_{t-1}(b_t),\\
\ell_{t, h_t} = h_t(b_t),
\end{array}} \right.
% \label{eq:}
% \end{split}
\end{equation}
as depicted in \cref{fig:maf parameters}(a).
$\ell_{t}$ and $\ell^{'}_{t}$ are the overall logits of two branches respectively. Taking $\ell_{t}$ as an example, its computation is
\begin{equation}
% \begin{split}
%   \ell_{t} &= \operatorname{H}_t(b_t) \\
%   &= [\ell_{t, \operatorname{H}_{t-1}}, \ell_{t, h_t}] \\
%   &=[\ell_{t, h_1},\cdots,\ell_{t, h_{t-1}},\ell_{t, h_t}],
%   \label{eq:l_t}
% \end{split}
\ell_{t} = \operatorname{H}_t(b_t) = [\operatorname{H}_{t-1}(b_t), \ell_{t, h_t}]  = [\ell_{t, \operatorname{H}_{t-1}}, \ell_{t, h_t}] 
\label{eq:l_t}
\end{equation}
where $[\cdot]$ is the concatenate operation and $\ell_{t} \in \mathbb{R}^{|C_1|+\cdots+|C_t|}$ covers all classes seen so far. 
The probabilistic output of $\ell_t$ is $p_t = \operatorname{Softmax}(\ell_t)$.
$\ell^{'}_{t}$ can be obtained in the similar way with $\bar{b}_{t-1}$ as input.

The residual fusion is applied to the corresponding logits between the two branches. Three fused logits are computed via
% and results in three kinds of outputs as
\begin{equation}
% \begin{split}
\left\{ {\begin{array}{*{20}{l}}
{{{{\rm{\bar \ell}}}_t} = \frac{1}{2}({\ell_t} + \ell_t^{'})},\\
{{{{\rm{\bar \ell}}}_{t,{\operatorname{H}_{t - 1}}}} = \frac{1}{2}({\ell_{t,{\operatorname{H}_{t - 1}}}} + \ell_{t,{\operatorname{H}_{t - 1}}}^{'})},\\
{{{{\rm{\bar \ell}}}_{t,{\operatorname{h}_t}}} = \frac{1}{2}({\ell_{t,{\operatorname{h}_t}}} + \ell_{t,{\operatorname{h}_t}}^{'})}.
\end{array}} \right.
\label{eq:fusion outputs3}
% \end{split}
\end{equation}
$\bar \ell_t$ is the fused logits of all categories till task $t$.
$\bar \ell_{t,\operatorname{H}_{t-1}}$ fuses the logits of old tasks.
$\bar \ell_{t,\operatorname{h}_{t}}$ is the logits for new categories at task $t$.
\subsection{Dynamic Residual Classifier}\label{sec:drc}

% The residual classifier (RC) described above still suffers from the growing storage and computation overhead by introducing the task-specific branch layers for CIL. 
The residual classifier (RC) still suffers from the growing storage and computation overhead as more task-specific branch layers are introduced under the CIL setting.
Dynamic Residual Classifier is proposed to handle this issue via the \emph{branch layer merging} in an iterative manner.
Assuming the task increment proceeds from $t-1$ to $t$, $\operatorname{B}^{'}_{t-2}$ and $\operatorname{B}_{t-1}$ are the two branch layers inherited from task $t-1$. As the branch layers are instantiated with the lightweight $1\times1$ convolutional layer without bias term, $\operatorname{B}^{'}_{t-2}$ and $\operatorname{B}_{t-1}$ are parameterized by the weight matrices $\omega_{\operatorname{B}^{'}_{t-2}}$ and $\omega_{\operatorname{B}_{t-1}}$ respectively.
A new branch layer $\operatorname{B}^{'}_{t-1}$ is obtained by merging $\operatorname{B}^{'}_{t-2}$ and $\operatorname{B}_{t-1}$ in the parameter space,
\begin{equation}
  {\omega_{\operatorname{B}_{t-1}^{'}}} = \frac{{\omega_{{\operatorname{B}_{t-1}}}} + {\omega_{\operatorname{B}_{t - 2}^{'}}}}{2}.
  \label{eq:weight}
\end{equation}
$\operatorname{B}^{'}_{t-1}$ is frozen to preserve the discriminative knowledge learned from previous tasks, as shown in \cref{fig:maf parameters}(b). 
The model-growing problem in RC is thus handled by DRC with only two branch layers, $\operatorname{B}^{'}_{t-1}$ and $\operatorname{B}_{t}$, included for each task.
Moreover, it is interesting to show that the final logit $\bar \ell_{t-1}$ of the previous model $\Phi_{t-1}$ is consistent with the output logit $\ell^{'}_{t,\operatorname{H}_{t-1}}$ of branch $\operatorname{B}^{'}_{t-1}$ at task $t$,
\begin{equation}
\begin{array}{l}
 \bar \ell_{t-1}= \frac{1}{2}({\operatorname{H}_{t - 1}}({\operatorname{B}_{t - 1}}(f)) + {\operatorname{H}_{t - 1}}(\operatorname{B}_{t - 2}^{'}(f)))\\
 = \frac{1}{2}({\operatorname{H}_{t - 1}}({\omega _{{\operatorname{B}_{t - 1}}}}\cdot f) + {\operatorname{H}_{t - 1}}({\omega _{\operatorname{B}_{t - 2}^{'}}}\cdot f))\\
  = {\omega _{{\operatorname{H}_{t - 1}}}}\cdot (\frac{1}{2}({\omega _{{\operatorname{B}_{t - 1}}}} + {\omega _{\operatorname{B}_{t - 2}^{'}}}) \cdot f)  + {{\kappa_{{\operatorname{H}_{t - 1}}}}}\\
 = {\omega _{{\operatorname{H}_{t - 1}}}}\cdot ({\omega_{\operatorname{B}_{t - 1}^{'}}} \cdot f ) + {{\kappa}_{{\operatorname{H}_{t - 1}}}}\\
 = {\operatorname{H}_{t - 1}}(\operatorname{B}_{t - 1}^{{'}}(f))\\
 = \ell^{'}_{t,\operatorname{H}_{t-1}},
\end{array} 
\end{equation}
% {\color{red}{
where ${\omega _{{\operatorname{H}_{t - 1}}}}$ and ${{\kappa_{{\operatorname{H}_{t - 1}}}}}$ are the weight and bias of the classifier ${\operatorname{H}_{t - 1}}$ respectively.
% }}
% Such an observation further demonstrates the effectiveness of the simple branch layer merging.
The effectiveness of simple branch layer merging can be further demonstrated by this observation.
The output logits of the DRC are also computed with the residual fusion mechanism as in \cref{eq:fusion outputs3}.
\begin{figure}[t]
  \centering
%   \fbox{\rule{0pt}{2in} \rule{0.9\linewidth}{0pt}}
  \includegraphics[width=0.8\linewidth]{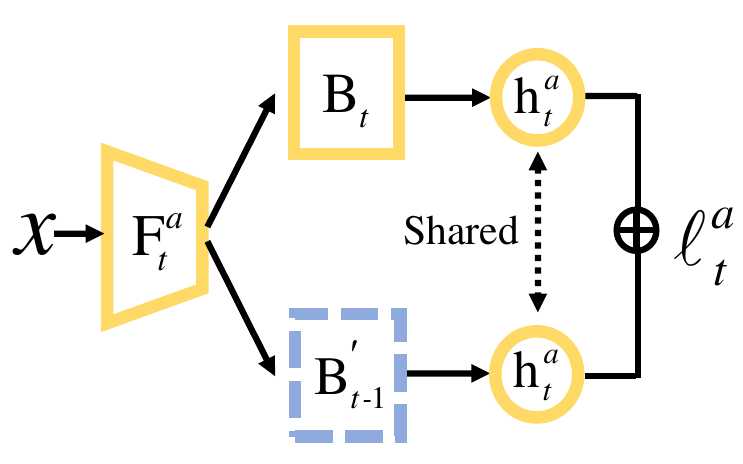}
   \caption{
   Model $\Phi_{t}^{a}$ with DRC for Adaptation.
   % Model Adaptation with DRC.
   % Model Adaptation with DRC on new data $D_t$.
   % The network architecture with DRC for model adaptation.
   % Illustration of the network architecture with DRC for model adaptation.
   }
   \label{fig:maf with drc}
\end{figure}

\subsection{CIL Pipelines with DRC}\label{sec:mafdrc}
% \subsection{Distillation Pipelines with DRC}\label{sec:mafdrc}
% \subsection{DRC under Different Pipelines}\label{sec:mafdrc}
% \subsection{Different Pipelines with DRC}\label{sec:mafdrc}

% \subsection{MAF with DRC}\label{sec:mafdrc}
% \subsection{Model Adaptation with DRC}\label{sec:madrc}

% Combining DRC with MAF does not change the pipeline. The overall procedure is very similar to \cref{sec:mafdrc}. The main difference comes from replacing the classifier $\operatorname{H}_t$ with $\operatorname{R}_t$.

% The proposed DRC is compatible with different CIL pipelines. 
The combinations between DRC and three CIL pipelines, the model adaptation and fusion (MAF), the model direct transfer (MDT), and the model expansion and compression (MEC), are presented.
The resulting methods, MAFDRC, MDTDRC, and MECDRC, are also compared.

% The combinations between the proposed DRC and three CIL pipelines respectively are presented.
% Combining the proposed DRC with three CIL pipelines, the model adaptation and fusion (MAF), the model direct transfer (MDT) and the model expansion and compression (MEC) respectively, results in three CIL methods, MAFDRC, MDTDRC and MECDRC.
% The combinations of the proposed DRC with three CIL pipelines, the model adaptation and fusion (), respectively are presented.
% Three pipelines of CIL 
% The proposed DRC is compatible with different CIL pipelines. 

\noindent \textbf{MAFDRC} \quad
The MAF pipeline consists of two successive stages, the model adaptation and then fusion, as shown in \cref{fig:frame}(c).
In the adaptation stage, a trainable model $\Phi^a_{t}$ with DRC is depicted as in \cref{fig:maf with drc}. 
Specifically, $\operatorname{B}_{t-1}^{'}$ is obtained via the branch layer merging as \cref{eq:weight} and fixed.
% initialized with the parameter averaging in \cref{eq:weight} and fixed.
$\ell_t^a$ is the residual fusion between the output logits of the two branches.
% The final logit of the model can be computed with \cref{eq:fusion outputs3} and denoted as $\ell_t^a$. 
% Its learning objective is still a cross-entropy as
With cross-entropy as the learning objective,
\begin{equation}
  \mathcal{L}_c^{a} =\sum\limits_{(x,y) \in {D_t}} {\operatorname{CE}(\ell_t^a,y)},
  %\frac{1}{{|{D_t}|}}
  \label{eq:adaptation loss}
\end{equation}
$\Phi_{t}^{a}$ is end-to-end optimized on current task $t$ with $D_t$ only.

In the model fusion stage, our aim is to integrate the knowledge from different models, i.e., $\Phi_{t-1}$ and $\Phi_{t}^{a}$, into a single $\Phi_t$ while reducing the impact of data imbalance.
$\Phi_t$ is instantiated with the model architecture as in \cref{fig:maf parameters}(b). $\operatorname{B}_{t-1}^{'}$ is from the branch layer merging and $\operatorname{B}_{t}$ is initialized with the corresponding branch of $\Phi_t^a$.
% The residual fused logits are computed as in \cref{eq:fusion outputs3}.
% In the model fusion stage, we want to integrate knowledge into a single model from the old and new model while reducing the impact of data imbalance.
% The new classifier can be constructed as ${\operatorname{R}_{t}} = \{ {\operatorname{B}_{t}},\operatorname{B}_{t-1}^{'}\}  \cup {\operatorname{H}_{t}}$. 
% The old model branch $\operatorname{B}_{t-1}^{'}$ is also initialized with parameter averaging and the new model branch $\operatorname{B}_{t}$ is initialized through the new model branch in the model adaptation step. 
% Two classification losses are combined to control the learning of the two branches,
% Two classification loss terms are proposed to guide the learning of the two branches,
The classification loss $\mathcal{L}_c^{branch}$ consists of two terms,
\begin{equation}
% \begin{split}
%   \mathcal{L}_c^{branch} =  \sum\limits_{(x,y) \in {D_t} \cup {M_t}} {\operatorname{CE}(\ell_t,y)} \\ + \sum\limits_{(x,y) \in  {M_t}} {\operatorname{CE}(\ell_{t,\operatorname{H}_{t-1}}^{'},y)}.
% \end{split}
\mathcal{L}_c^{branch} =  \sum\limits_{(x,y) \in {D_t} \cup {M_t}} {\operatorname{CE}(\ell_t,y)} + \sum\limits_{(x,y) \in  {M_t}} {\operatorname{CE}(\ell_{t,\operatorname{H}_{t-1}}^{'},y)}.
\label{eq:branch loss}
\end{equation}
They guide the learning of the two branches and encourage parameter specialization for old and new classes respectively.
% These two terms encourage parameter specialization for old and new classes respectively.
% A fusion loss item ${L_c^{fusion}}$ is also used to optimize for all the classes, with $\bar \ell_t$ computed as in \cref{eq:fusion outputs3},
${L_c^{fusion}}$ is proposed to optimize the model over all the classes across tasks,
% with $\bar \ell_t$ computed as in \cref{eq:fusion outputs3},
with the fused logit $\bar \ell_t$ from \cref{eq:fusion outputs3},
\begin{equation}
\begin{split}
  {\mathcal{L}_c^{fusion}} =\sum\limits_{(x,y) \in {D_t} \cup {M_t}} {\operatorname{CE}(\bar \ell_t,y)}.
  %  \frac{1}{{|{D_t}| + |{M_t}|}}
  \label{eq:fusion loss}
\end{split}
\end{equation}
Moreover, the discriminative knowledge from $\Phi_{t}^a$ and $\Phi_{t-1}$ are distilled into the final model $\Phi_{t}$ with the loss,
% . The knowledge distillation loss becomes
\begin{equation}
%\begin{split}
\begin{split}
    \mathcal{L}_{distil}= \sum\limits_{(x,y) \in {D_t} \cup {M_t}} \operatorname{KL}(p_{t - 1}||\bar p_{t,\operatorname{H}_{t-1}})  + \operatorname{KL}(p_t^a||\bar p_{t,\operatorname{h}_t}),
    % \frac{1}{{|{D_t}| + |{M_t}|}}
  \label{eq:new kd}
\end{split}
\end{equation}
where $\bar p_{t,\operatorname{H}_{t-1}}$ can be represented by $\operatorname{S}(\bar \ell_{t,\operatorname{H}_{t-1}})$, and $\bar p_{t,\operatorname{h}_{t}}$ can be represented by $\operatorname{S}(\bar \ell_{t,\operatorname{h}_{t}})$).
In the model fusion step, $\Phi_t$ is optimized on the overall loss,
\begin{equation}
  {\mathcal{L}_{all}} = (1 - \alpha ){\mathcal{L}_c^{fusion}} + \alpha {\mathcal{L}_c^{branch} + \beta {\mathcal{L}_{distil}}},
  \label{eq:sum}
\end{equation}
with balancing hyper-parameters $0 \leq \alpha \leq 1$ and $\beta$.
% to keep the balance.

%-------------------------------------------------------------------------
% \subsection{Other Pipeline with DRC} \label{sec:othdrc}

% In this section, we will introduce how our dynamic residual classifier can be used on other pipelines. 

\noindent \textbf{MDTDRC} \quad
In the MDT pipeline, distillation loss is used to directly transfer the knowledge of the previous task from $\Phi_{t-1}$ (the teacher) to $\Phi_t$ (the student) and prevent the representations of previous data from drifting too much during new task learning, as illustrated in \cref{fig:frame}(a).
% % In {\bf model direct transfer} pipeline, 
% In the MDT pipeline, a new model $\Phi_t$ is initialized by the old one $\Phi_{t-1}$ and then directly fine-tuned with the classification and distillation losses on $M_t$ and $D_t$, as illustrated in \cref{fig:frame}(a).
The proposed DRC is enabled by introducing the branch layers, $\operatorname{B}_{t}$ and $\operatorname{B}_{t-1}^{'}$. The parameters of $\operatorname{B}_{t-1}^{'}$ are obtained via \cref{eq:weight} and $\operatorname{B}_{t}$ for the new task is random initialized. 
% Different from MAFDRC, this method is learned within a sin
% that only uses a classification loss under the constraints by the distillation of old model (similar to iCaRL). 
% The two branches $\operatorname{B}_{t}$ and $\operatorname{B}_{t-1}^{'}$ are also added to form a new classifier ${\operatorname{R}_{t}} = \{ {\operatorname{B}_{t}},\operatorname{B}_{t-1}^{'}\}  \cup {\operatorname{H}_{t}}$ between the feature extractor $\operatorname{F}_{t}$ and the classifier $\operatorname{H}_{t}$ for each step.
% The parameters of $\operatorname{B}_{t-1}^{'}$ are also obtained by \cref{eq:weight} and the new model branch $\operatorname{B}_{t}$ is initialized randomly.
% The outputs of the model are similar to our pipeline, including $\ell_t$, $\ell_{t,\operatorname{H}_{t-1}}^{'}$ and $\bar \ell_t$.
% Two classification losses (similar to \cref{eq:branch loss}) are used to control the learning of the two branches and a fusion loss (similar to \cref{eq:fusion loss}) is to optimize all the classes. 
% The fused outputs $\bar \ell_t$ are used as final results both in the training and test phase. 

\noindent \textbf{MECDRC} \quad
In the model expansion stage, the previous model is frozen and expanded with a new trainable one by concatenating their feature representations.
The resulting larger model is then trained with $D_t$ and $M_t$, as shown in \cref{fig:frame}(b).
The proposed DRC is also introduced at this stage to boost the performance of the large model across tasks. % How to add DRC to this stage? details?
In the mode compression stage, the expanded model is distilled into the final model with a smaller size.

\noindent \textbf{Comparisons among Three Pipelines} \quad
MDT usually impose a challenge to find the balance between learning novel classes and preserving old knowledge simultaneously within a single model. 
Model expansion can achieve better balance at the price of storage and computation overhead, while model compression may neutralize the improvements from expansion.
% In the MAF pipeline, 
In MAF, $\Phi_t^a$ is optimized with only new data $D_t$ in the first stage. $\Phi_{t-1}$ preserves the knowledge of previous tasks. 
% In the next stage, 
The complementary knowledge in $\Phi_t^a$ and $\Phi_{t-1}$ is then fused into $\Phi_t$ by distillation to improve performance on both old and new tasks.
% Therefore, the two stages of MAF are more integrated than those of MEC.
Moreover, DRC is integrated into both stages of MAF rather than a single stage of MEC or MDT.
% Moreover, DRC is integrated into both stages of MAF. 
To this end, MAFDRC is chosen as our main method.

% Compared with the three pipelines, MAF is a more consistent and integrated pipeline. First, the model training in each MAF stage is optimized for one specific purpose. No effort has been made in balancing the contradicting objectives as in MDT. Second, the $\Phi^a_t$ learned in the adaptation stage is the source of discriminative knowledge of the new task in model fusion. The two stages of MAF are complementary and integrated, rather than neutralizing the significance of each other as in MEC.

%------------------------------------------------------------------------
\section{Experiments}
\label{sec:exper}

\begin{table*}[t]
\centering
% \scalebox{1.0}
% {
\begin{tabular}{c|cccccc|cccc|cc}
\hline
\multirow{3}{*}{Methods} & \multicolumn{6}{c|}{ImageNet100 B0}                                                        & \multicolumn{4}{c|}{ImageNet100 B50}                        & \multicolumn{2}{c}{ImageNet1000} \\ \cline{2-13} 
                         & \multicolumn{2}{c}{5 steps} & \multicolumn{2}{c}{10 steps} & \multicolumn{2}{c|}{20 steps} & \multicolumn{2}{c}{5 steps} & \multicolumn{2}{c|}{10 steps} & \multicolumn{2}{c}{10 steps}     \\ \cline{2-13} 
                         & Avg          & Last         & Avg          & Last          & Avg           & Last          & Avg          & Last         & Avg           & Last          & Avg            & Last            \\ \hline
                
iCaRL~\cite{icarl}     & 74.87                                 & 63.36                                     & 70.35                                 & 55.78                                     & 67.80                                 & 51.78  & 64.69                                 & 54.46                                 & 57.92                                 & 50.52 & 54.15                                 & 36.25       \\
BiC~\cite{wu2019large}  & 77.11                                 & 67.10                                     & 70.98                                 & 52.00                                     & 63.79                                 & 41.70   & 68.51                                 & 54.36                                 & 60.73                                 & 43.04        & 61.66                                     & 41.30     \\
WA~\cite{wa}        & 77.59                                 & 68.36                                     & 73.59                                 & 60.78                                     & 68.81                                 & 57.16     & 68.49                                 & 59.74                                 & 62.10                                 & 54.42   & 59.23                                 & 40.92       \\
PODNet~\cite{podnet}    & 76.73                                 & 64.90                                     & 70.13                                 & 53.30                                     & 62.78                                 & 47.10    & 78.41                                 & 69.18                                 & 75.97                                 & 66.50  & - & - \\ 
DER w/o p~\cite{der} & {\color[HTML]{3166FF} \textbf{81.03}}                                 & {\color[HTML]{3166FF} \textbf{74.44}}                                             & {\color[HTML]{3166FF} \textbf{78.30}}                                 & {\color[HTML]{3166FF} \textbf{70.40}}                                                      & {\color[HTML]{FE0000} \textbf{78.22}}                                 & {\color[HTML]{FE0000} \textbf{71.40}}      & {\color[HTML]{3166FF} \textbf{80.30}} & {\color[HTML]{3166FF} \textbf{74.28}} & {\color[HTML]{FE0000} \textbf{78.58}} & {\color[HTML]{FE0000} \textbf{71.66}}    & 67.41                                 & {\color[HTML]{3166FF} \textbf{58.56}} \\

% 3EF~\cite{choi2021dual} & -   & -   & -    & -   & 77.62   & 68.78   & 93.66    & 89.32  & -    & -    & -    & -   \\   

% DyTox~\cite{dytox} & 74.79   & 68.20   & 92.12    & 89.38   & 71.85   & 57.94   & 90.72    & 83.52  & -    & -    & -    & -   \\

FOSTER B4~\cite{foster} & 79.59 & 72.58         & 76.54 & 67.08               & 74.21 & 62.16             & 79.93                                 & 72.48                                 & 76.27                                 & 67.04                   & {\color[HTML]{3166FF} \textbf{68.34}} & 58.53         \\
FOSTER~\cite{foster}    & 78.38                                 & 71.38                                 & 76.22                                 & 66.70                                  & 73.95                                 & 62.42       & 79.56                                 & 71.18                                 & 75.79                                 & 66.90    & - & -  \\ \hline
MAFDRC    & {\color[HTML]{FE0000} \textbf{82.22}} & {\color[HTML]{FE0000} \textbf{76.01}} & {\color[HTML]{FE0000} \textbf{79.66}} & {\color[HTML]{FE0000} \textbf{70.41}} & {\color[HTML]{3166FF} \textbf{75.21}} & {\color[HTML]{3166FF} \textbf{63.59}}   & {\color[HTML]{FE0000} \textbf{81.37}} & {\color[HTML]{FE0000} \textbf{74.86}} & {\color[HTML]{3166FF} \textbf{77.95}} & {\color[HTML]{3166FF} \textbf{71.26}}  & {\color[HTML]{FE0000} \textbf{69.37}} & {\color[HTML]{FE0000} \textbf{59.59}}\\ \hline
\end{tabular}
\caption{
% {\color{red}{
Results on ImageNet100 B0, B50 and ImageNet1000 B0 settings.
% }}
% Results on ImageNet.
DER w/o P means DER without pruning. 
FOSTER B4 means the model before feature compression.
% FOSTER B4 means the model performance before feature compression.
}
% , Ours vs. state of the art. 
% DER w/o P means DER without pruning. FOSTER B4 means the model performance before feature compression.}
\label{tab:imagenet}
\end{table*}

\begin{figure*}[t]
\centering
  % \begin{subfigure}{0.32\linewidth}
  %   % \fbox{\rule{0pt}{2in} \rule{.9\linewidth}{0pt}}
  %   \includegraphics[width=1\linewidth]{figs/Imagenet100_results5.pdf}
  %   \caption{ImageNet100 5 steps.}
  %   \label{fig:imagenet100-b0-5}
  % \end{subfigure}
  % \hfill
  \begin{subfigure}{0.32\linewidth}
    % \fbox{\rule{0pt}{2in} \rule{.9\linewidth}{0pt}}
    \includegraphics[width=1\linewidth]{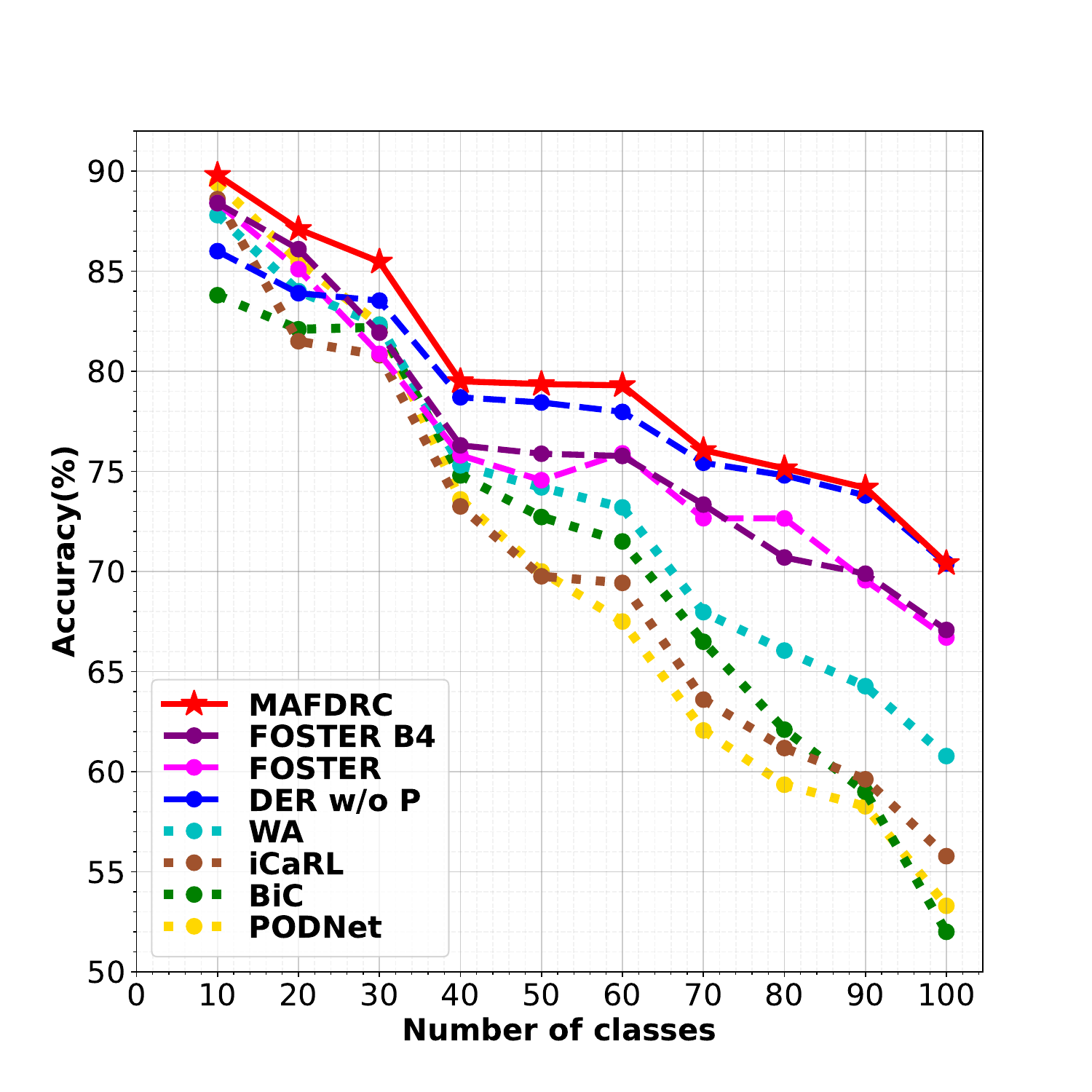}
    \caption{ImageNet100 B0 10 steps}
    \label{fig:imagenet100-b0-10}
  \end{subfigure}
  \hfill
  % \begin{subfigure}{0.32\linewidth}
  %   % \fbox{\rule{0pt}{2in} \rule{.9\linewidth}{0pt}}
  %   \includegraphics[width=1\linewidth]{figs/Imagenet100_results20.pdf}
  %   \caption{B0 20 steps}
  %   \label{fig:imagenet100-b0-20}
  % \end{subfigure}
  % \hfill
  \begin{subfigure}{0.32\linewidth}
    % \fbox{\rule{0pt}{2in} \rule{.9\linewidth}{0pt}}
    \includegraphics[width=1\linewidth]{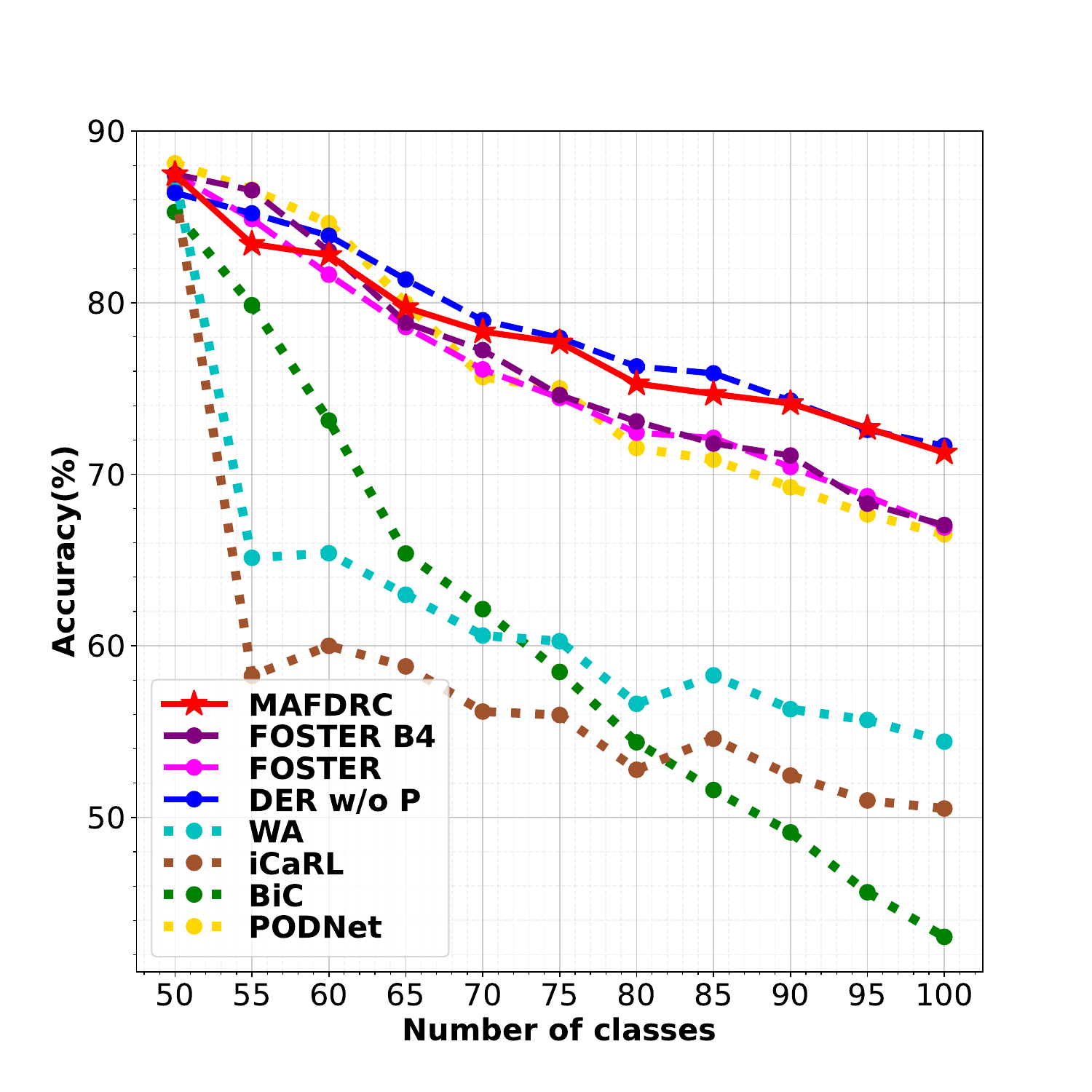}
    \caption{ImageNet100 B50 10 steps}
    \label{fig:imagenet100-b50-10}
  \end{subfigure}
  \hfill
  \begin{subfigure}{0.32\linewidth}
    % \fbox{\rule{0pt}{2in} \rule{.9\linewidth}{0pt}}
    \includegraphics[width=1\linewidth]{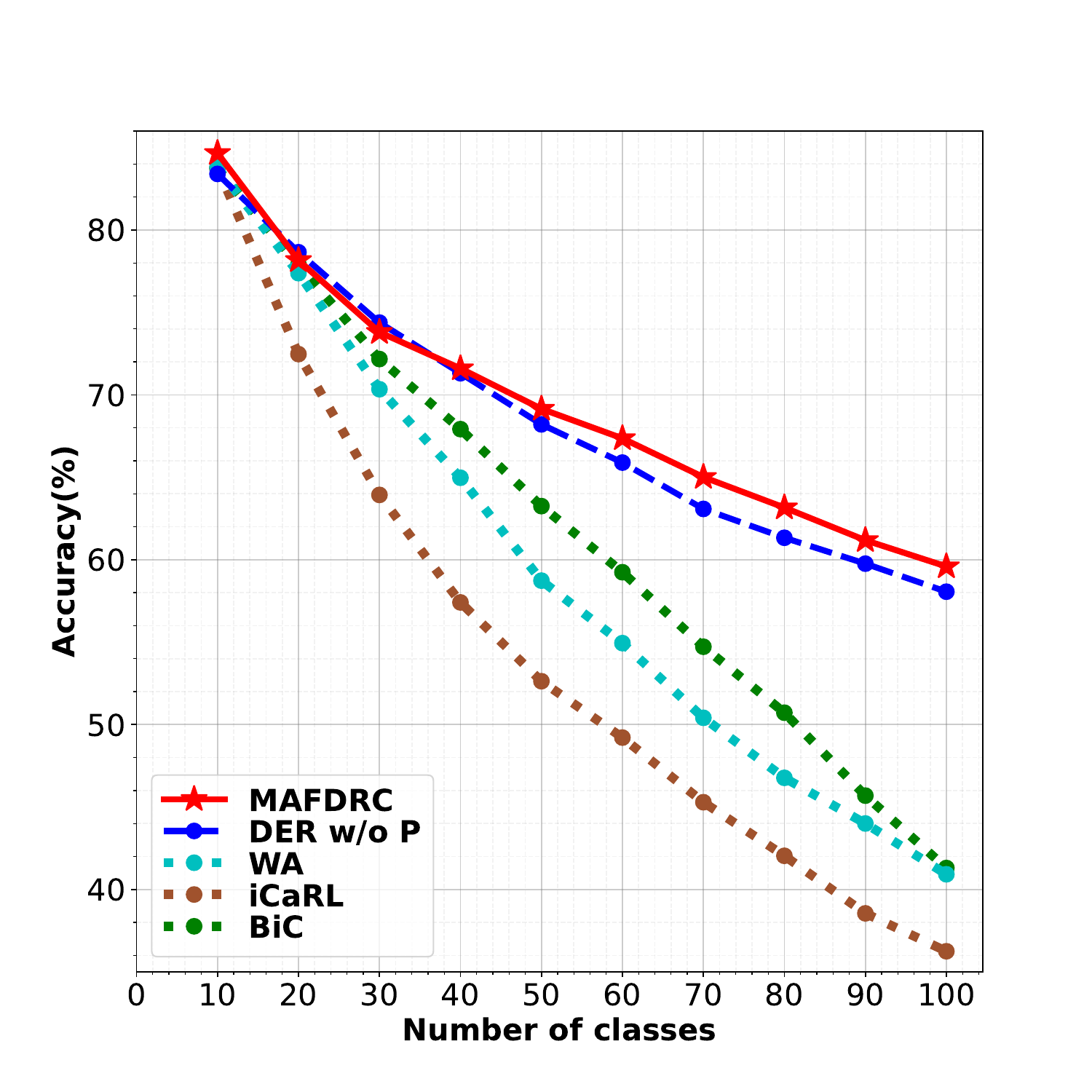}
    \caption{ImageNet1000 B0 10 steps}
    \label{fig:imagenet1000-b0-10}
  \end{subfigure}
  \hfill
  \caption{
  Incremental Accuracy on ImageNet.
  % Incremental Accuracy on ImageNet100.
  The top-1 accuracy (\%) after learning each task is shown.
  }
  % Left is evaluated with 5 steps, middle with 10 steps, and right with 20 steps.}
  \label{fig:imagenet}
\end{figure*}

%-------------------------------------------------------------------------
% \subsection{Experiment Setup and Implementation Details}\label{sec:DES}
\subsection{Experimental Setup and Details}\label{sec:DES}

\noindent {\bf Datasets } \quad
% { \bf Datasets and Protocols} \quad
% { \bf Datasets and Benchmarks} \quad
ImageNet1000~\cite{deng2009imagenet} is a large-scale dataset with 1,000 classes that includes 1.28 million images for training and 50,000 images for validation. ImageNet100~\cite{deng2009imagenet} is made up of 100 randomly selected classes from ImageNet1000. CIFAR100~\cite{krizhevsky2009learning} consists of 100 classes and each class has 600 images, of which 500 are used as the training set and 100 are used as the test set.  

\noindent {\bf Protocols} \quad
For ImageNet100 and CIFAR100, we validate the proposed method in two widely used CIL protocols:
(1) {\bf B0} (base 0)~\cite{icarl}: In this protocol, a model is gradually trained on 5 steps (20 new classes per step), 10 steps (10 new classes per step) and 20 steps (5 new classes per step) with the fixed memory size of 2,000 exemplars.
(2) {\bf B50} (base 50)~\cite{hou2019learning}: A model is trained first on 50 classes. The remaining 50 classes are used for continual learning with 5, 10 classes per step. The memory size is fixed to 20 exemplars per class.
For ImageNet1000, we evaluate our method on the protocol~\cite{icarl} where a model is trained on all 1,000 classes with 100 classes per step (10 steps in total). The fixed memory size is 20,000 exemplars.
We also carried out the LT-CIL experiments on CIFAR100 and ImageNet100 following the protocol~\cite{liu2022long}.
% The LT-CIL setting on the CIFAR100 and ImageNet100 is proposed in \cite{liu2022long}
% The LT-CIL setting~\cite{liu2022long} is built on the CIFAR100 and ImageNet100.
Two classification accuracy are reported. "Avg" is the average accuracy over incremental steps. "Last" is the accuracy of the last step.

% For the CIFAR100/ImageNet100 B0 benchmark\footnote{
% Refer to Sec.~\ref{supp_sec:B50-results} in supplementary for B50 results .
% % The results refer to Sec.~\ref{supp_sec:B50-results} in the supplementarygg
% }, we test our methods on 5 steps (20 new classes per step), 10 steps (10 new classes per step), 20 steps (5 new classes per step) with the fixed memory size of 2,000 exemplars which follow the protocol proposed in ~\cite{icarl}. 
% % For benchmark {\bf CIFAR100/ImageNet100 B50}\footnote{
% % Refer to Sec.~\ref{supp_sec:B50-results} in supplementary.
% % % The results refer to Sec.~\ref{supp_sec:B50-results} in the supplementarygg
% % }, we first train half of 100 classes at the base learning stage. Then we train the rest 50 classes with 5, 10 classes per step. Different from the above protocol, models are allowed to store 20 exemplars for each classs and no more than 2,000 exemplars.
% For ImageNet1000, we evaluate our method on the protocol~\cite{icarl}, that adds 100 classes per step (10 steps in total) with a fixed memory size of 20,000 exemplars. 
% We compare the top-1 and top-5 average incremental accuracy and the last step accuracy on ImageNet datasets. We only use the top-1 average incremental accuracy and the last step accuracy on CIFAR100 datasets. 
% The "Avg" accuracy is the average accuracies of all previous tasks after each step as defined in ~\cite{icarl}. The "Last" accuracy is the final accuracy after the last step.

\noindent {\bf Implementation Details} \quad
% 1.based on pytorch, except the partial results of ImageNet1000 is it necessary?
Our method is implemented with PyTorch~\cite{grossberg2012studies} and PyCIL~\cite{DBLP}.
% For a fair comparison, 
The standard 18-layer ResNet~\cite{he2016deep} is used as the backbone feature extractor for ImageNet. For CIFAR100, a modified 32-layer ResNet~\cite{icarl} is used instead.
% {\color{red}{
In our experiments, the MDT setting follows~\cite{icarl}, that of MEC is from~\cite{foster}, and the MAF one is based on~\cite{hou2018lifelong} in all experiments.
% }}
Integrating the proposed DRC with the MAF pipeline results in our main method, MAFDRC.
For different CIL settings, similar optimization is used.
% our method is optimized under similar settings.
In model adaptation, the SGD optimizer with a momentum of 0.9 is used for training 70 epochs in total. The initial learning rate is set to 0.1 and gradually reduces to zero with a cosine annealing scheduler. In model fusion, the above settings are followed, but training 130 epochs in total.
For data augmentation, we follow the practice\footnote{\url{https://github.com/G-U-N/ECCV22-FOSTER}.} in FOSTER~\cite{foster}, where the AutoAugment~\cite{cubuk2019autoaugment} is used along with the common random cropping and horizontal flip.
For a fair comparison, the reported CIL results of all competitors in this work are reproduced with such an augmentation.\footnote{We also conduct experiments with conventional data augmentation and report such results in the Supplementary Material.}
In the LT-CIL scenario, our experimental setting is exactly the same as in~\cite{liu2022long}.
The balancing hyperparameters ${\alpha}$ and $\beta$ in \cref{eq:sum} are set to 0.2 and 4 respectively via cross-validation.
\begin{table*}[t]
\centering
% \scalebox{1.0}
% {
\begin{tabular}{c|cccccc|cccc}
\hline
\multirow{3}{*}{Methods} & \multicolumn{6}{c|}{CIFAR100 B0}                                                        & \multicolumn{4}{c}{CIFAR100 B50}                        \\ \cline{2-11} 
                         & \multicolumn{2}{c}{5 steps} & \multicolumn{2}{c}{10 steps} & \multicolumn{2}{c|}{20 steps} & \multicolumn{2}{c}{5 steps} & \multicolumn{2}{c}{10 steps} \\ \cline{2-11} 
                         & Avg          & Last         & Avg          & Last          & Avg           & Last          & Avg          & Last         & Avg          & Last          \\ \hline

iCaRL ~\cite{icarl}     & 69.29                                 & 57.03                                 & 68.37                                 & 53.04                                 & 67.43                                 & 49.65                   & 62.21                                 & 53.63                                 & 53.65                                 & 47.18                         \\
BiC ~\cite{wu2019large}       & 68.66                                 & 58.22                                 & 67.75                                 & 53.31                                 & 65.41                                 & 47.12                     & 63.92                                 & 54.18                                 & 59.68                                 & 48.04                         \\
WA ~\cite{wa}        & 72.09                                 & 61.49                                 & 70.88                                 & 56.74                                 & 68.10                                 & 49.60                         & 67.30                                 & 59.37                                 & 61.86                                 & 50.86                      \\
PODNet ~\cite{podnet}    & 69.32                                 & 57.75                                 & 63.17                                 & 47.49                                 & 58.26                                 & 40.62                       & 70.40                                 & 62.49                                 & 69.20 & 60.14                    \\ 

% 3EF~\cite{wang3ef} & 73.05   & 62.48      & 72.93   & 61.45      & 71.69    & 57.06  & 71.58 & 64.54 & {\color[HTML]{3166FF} \textbf{71.70}} & 61.19\\   

% DyTox~\cite{dytox} & 73.02   & 61.31      & 71.50    & 57.76  & 68.86       & 51.47 & & & &  \\
DER w/o P ~\cite{der} & {\color[HTML]{FE0000} \textbf{75.83}}                                 & {\color[HTML]{FE0000} \textbf{68.95}}                                 & {\color[HTML]{FE0000} \textbf{75.71}}                                 & {\color[HTML]{FE0000} \textbf{65.85}}                                 & {\color[HTML]{FE0000} \textbf{74.04}}                                 & {\color[HTML]{FE0000} \textbf{62.53}}   & {\color[HTML]{FE0000} \textbf{72.95}} & {\color[HTML]{FE0000} \textbf{68.06}} & {\color[HTML]{FE0000} \textbf{72.50}} & {\color[HTML]{FE0000} \textbf{67.37}} \\
FOSTER B4 ~\cite{foster} & 74.53 & 65.31 & 73.13 & 61.81 & 70.64 & 56.84                     & 71.31                                 & 64.66                                 & 68.90                                 & 61.41               \\
FOSTER ~\cite{foster}    & 72.46                                 & 63.35                                 & 71.80                                 & 60.15                                 & 69.56                                 & 56.50                           & 70.09                                 & 63.63                                 & 68.05                                 & 60.71                      \\ \hline
MAFDRC      & {\color[HTML]{3166FF} \textbf{74.87}} & {\color[HTML]{3166FF} \textbf{66.45}} & {\color[HTML]{3166FF} \textbf{73.97}} & {\color[HTML]{3166FF} \textbf{62.04}} & {\color[HTML]{3166FF} \textbf{71.75}} & {\color[HTML]{3166FF} \textbf{57.65}}     & {\color[HTML]{3166FF} \textbf{71.65}} & {\color[HTML]{3166FF} \textbf{65.09}} & {\color[HTML]{3166FF} \textbf{70.21}}                                 & {\color[HTML]{3166FF} \textbf{62.20}}        \\ \hline

\end{tabular}
\caption{
% {\color{red}{
Results on CIFAR100 B0, B50 setting.
% }}
% Results on CIFAR100.
}
% , Ours vs. state of the art. 
% DER w/o P means DER without pruning. FOSTER B4 means the model performance before feature compression.}
\label{tab:cifar}
\end{table*}

\begin{figure*}[t]
  \centering
  % \begin{subfigure}{0.32\linewidth}
  %   % \fbox{\rule{0pt}{2in} \rule{.9\linewidth}{0pt}}
  %   \includegraphics[width=1\linewidth]{figs/cifar_results5.pdf}
  %   \caption{CIFAR100 5 steps.}
  %   \label{fig:cifar-5}
  % \end{subfigure}
  % \hfill
  \begin{subfigure}{0.32\linewidth}
    % \fbox{\rule{0pt}{2in} \rule{.9\linewidth}{0pt}}
    \includegraphics[width=1\linewidth]{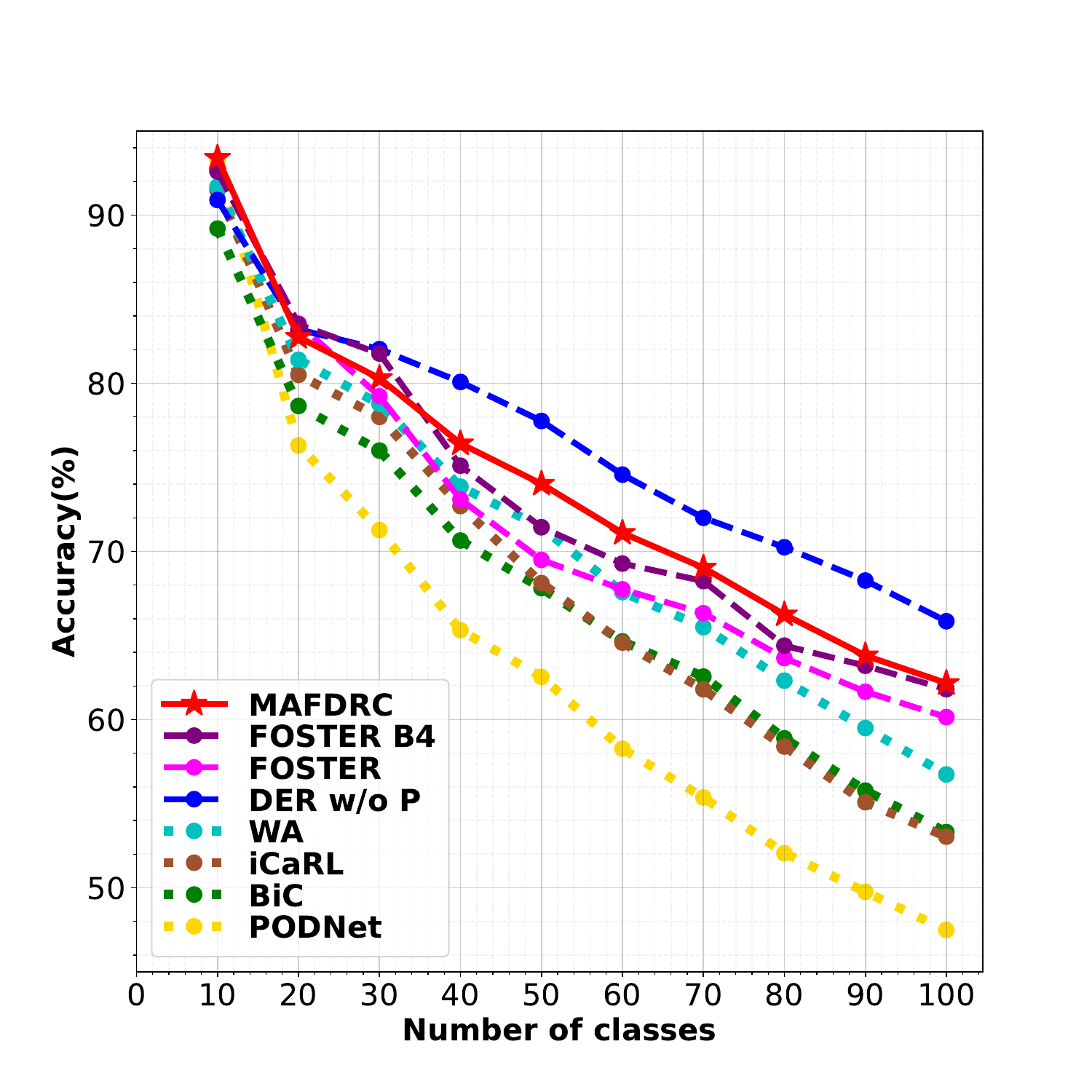}
    \caption{B0 10 steps}
    \label{fig:cifar-b0-10}
  \end{subfigure}
  \hfill
  \begin{subfigure}{0.32\linewidth}
    % \fbox{\rule{0pt}{2in} \rule{.9\linewidth}{0pt}}
    \includegraphics[width=1\linewidth]{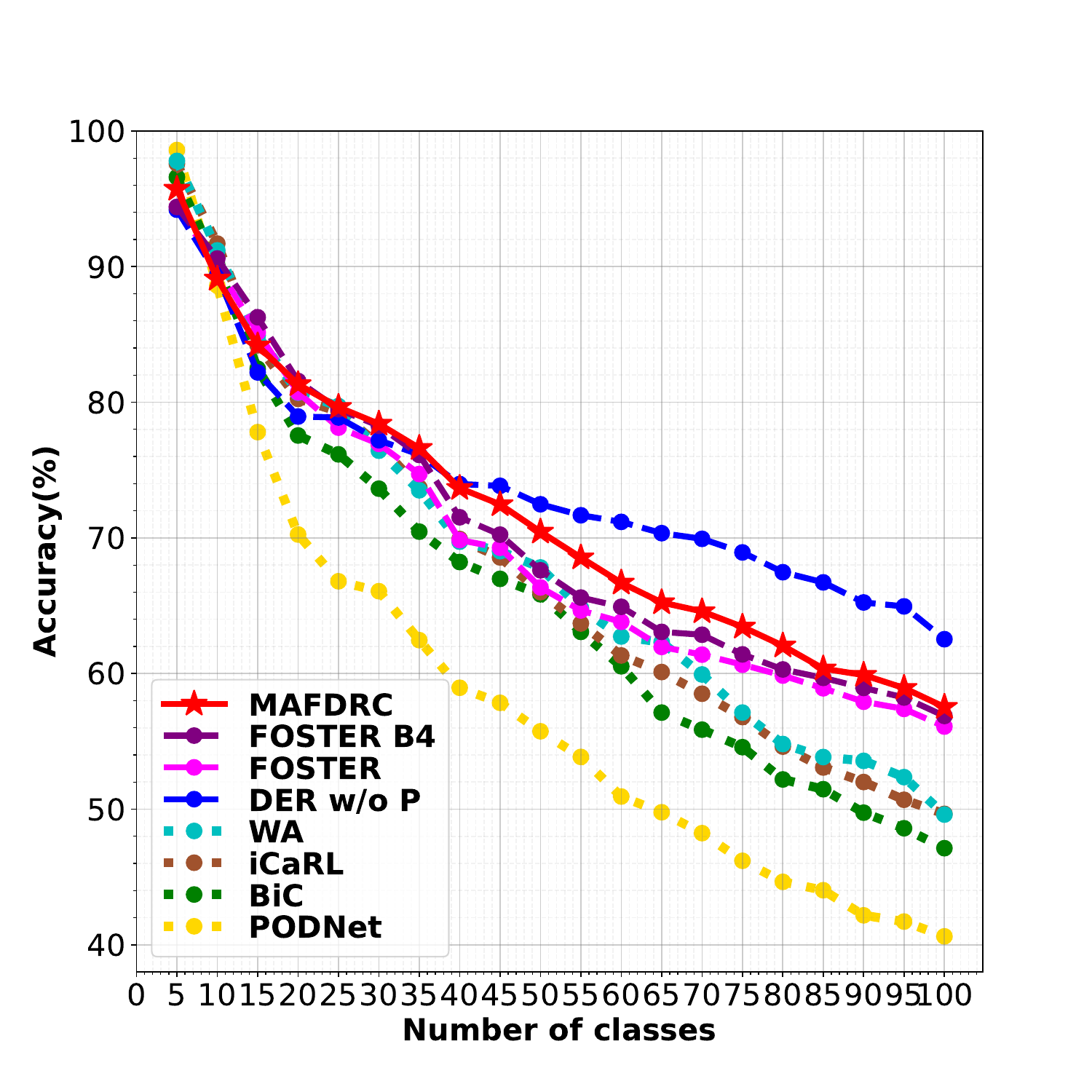}
    \caption{B0 20 steps}
    \label{fig:cifar-b0-20}
  \end{subfigure}
  \begin{subfigure}{0.32\linewidth}
    % \fbox{\rule{0pt}{2in} \rule{.9\linewidth}{0pt}}
    \includegraphics[width=1\linewidth]{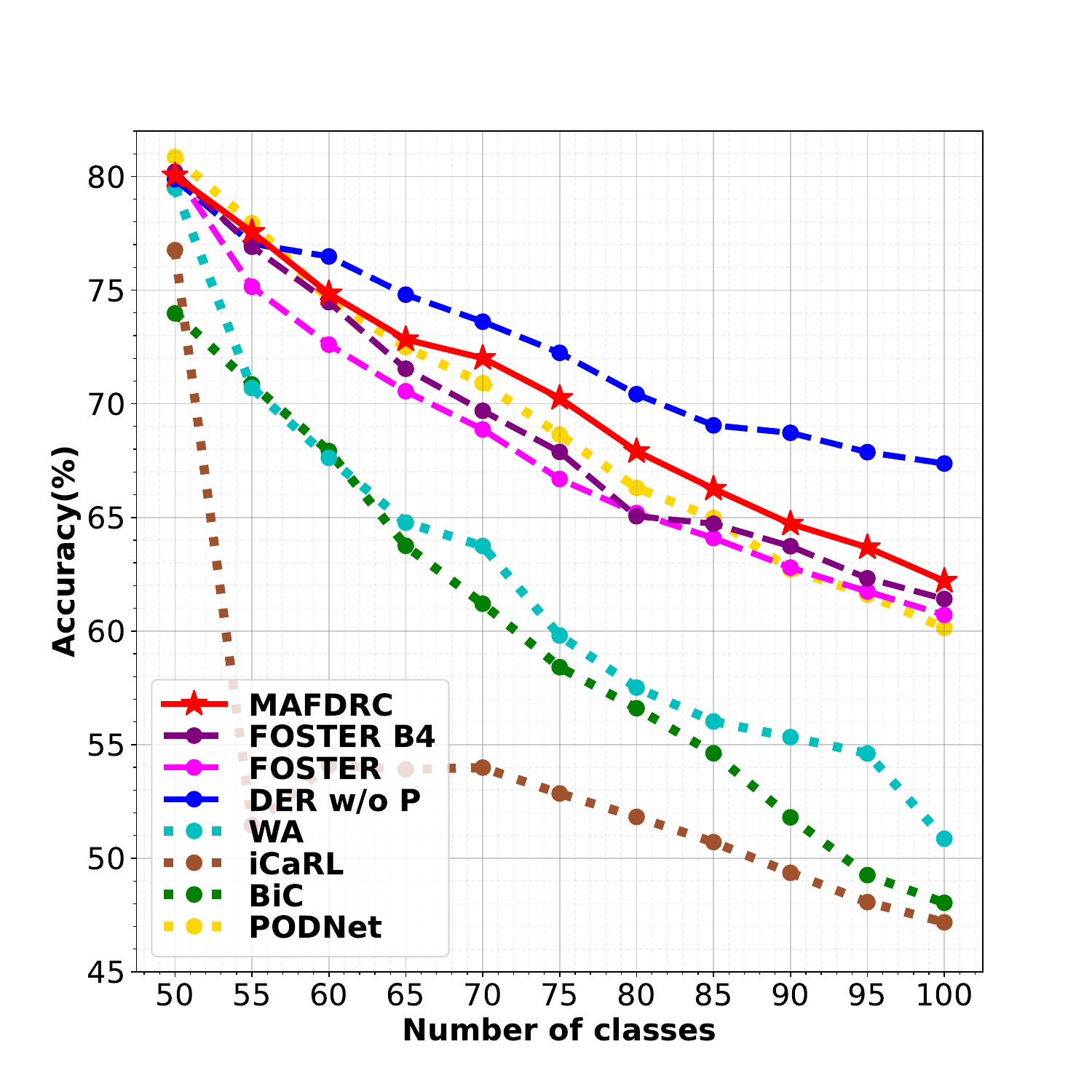}
    \caption{B50 10 steps}
    \label{fig:cifar-b50-10}
  \end{subfigure}
  \hfill
  \caption{ Incremental Accuracy on CIFAR100. The top-1 accuracy (\%) after learning each task is shown.}
  % Left is evaluated with 5 steps, middle with 10 steps, and right with 20 steps.}
  % These are averaged over 3 runs.}
  \label{fig:cifar}
\end{figure*}

\begin{table}[t]
\centering
\scalebox{0.9}
{
\begin{tabular}{cccccc}
\hline
\multicolumn{2}{c}{}                             & \multicolumn{2}{c}{CIFAR100}                                & \multicolumn{2}{c}{ImageNet100} \\ \cline{3-6} 
\multicolumn{2}{c}{\multirow{-2}{*}{Methods}}    & 5 steps                      & 10 steps                     & 5 steps        & 10 steps       \\ \hline
                                  & EEIL         & 38.46                        & 37.50                        & 50.68          & 50.63          \\
                                  & +Two Stage\cite{liu2022long} & 38.97                        & 37.58                        & 51.36          & 50.74          \\ \cline{2-6} 
                                  & LUCIR         & 42.69                        & 42.15                        & 52.91          & 52.80          \\
                                  & +Two Stage\cite{liu2022long} & \color[HTML]{3166FF} \textbf{45.88} & \color[HTML]{3166FF} \textbf{45.73} & 54.22          & 55.41          \\ \cline{2-6} 
                                  & PODNet       & 44.07                        & 43.96                        & 58.78          & 58.94          \\
                                  & +Two Stage\cite{liu2022long} & 44.38                        & 44.35                        & \color[HTML]{3166FF} \textbf{58.82}          & \color[HTML]{3166FF} \textbf{59.09}          \\ \cline{2-6} 
                                  
                                  & MAF       & 35.92 & 33.70 & 46.62             & 35.83             \\
                                  % & DER$^*$       & 50.64                        & 50.00                        & -          & -          \\
                                  % & +(Two Stage)$^*$ & -                        & -                        & \color[HTML]{FE0000} \textbf{-}          & -          \\ \cline{2-6} 
\multirow{-8}{*}{\rotatebox{90}{Ordered LT-CIL}}  & MAFDRC       & \color[HTML]{FE0000} \textbf{53.13} & \color[HTML]{FE0000} \textbf{49.01} & \color[HTML]{FE0000} \textbf{60.69}              & \color[HTML]{FE0000} \textbf{59.65}              \\ \hline

                                  & EEIL         & 31.91                        & 32.44                        & 42.87          & 43.72          \\
                                  & +Two Stage\cite{liu2022long} & 34.19                        & 33.70                        & 49.31          & 48.26          \\ \cline{2-6} 
                                  & LUCIR        & 35.09                        & 34.59                        & 45.80          & 46.52          \\
                                  & +Two Stage\cite{liu2022long} & \color[HTML]{3166FF} \textbf{39.40} & \color[HTML]{3166FF} \textbf{39.00} & \color[HTML]{3166FF} \textbf{52.08}          & 51.91          \\ \cline{2-6} 
                                  & PODNet       & 34.64                        & 34.84                        & 49.69          & 51.05          \\
                                  & +Two Stage\cite{liu2022long} & 36.37                        & 37.03                        & 51.55          & \color[HTML]{3166FF} \textbf{52.60}          \\ \cline{2-6} 

                                  & MAF       & 31.63 & 30.18 & 41.53             & 39.92             \\
                                  % & DER       & -                        & 29.54                        & -          & -          \\
                                  % & +(Two Stage) & -                        & 33.54                        & \color[HTML]{FE0000} \textbf{-}          & -          \\ \cline{2-6} 
                                  % & DER$^*$       & 30.00                        & 32.00                        & -          & -          \\
                                  % & +(Two Stage)$^*$ & -                        & -                        & \color[HTML]{FE0000} \textbf{-}          & -          \\ \cline{2-6}
\multirow{-8}{*}{\rotatebox{90}{Shuffled LT-CIL}} & MAFDRC       & \color[HTML]{FE0000} \textbf{41.41} & \color[HTML]{FE0000} \textbf{41.84} & \color[HTML]{FE0000} \textbf{52.49}              & \color[HTML]{FE0000} \textbf{56.35}              \\ \hline
\end{tabular}
}
\caption{
% {\color{red}{
% LT-CIL results on B50 setting with average accuracy reported. 
% }}
% \textcolor{red}{
LT-CIL~\cite{liu2022long}
% } 
results with average accuracy reported. 
% Results on average accuracy under the LT-CIL setting.
% The results under the LT-CIL setting. 
% Two stage means that decouples representation learning from classifier learning as in ~\cite{liu2022long}.
}
\label{tab:LT-CIL}
\end{table}
% 这个地方的DER因为Two Stage的方法无法准确复现，是不是可以在文中说一下结果就好。

% \subsection{Quantitative Results}\label{sec:imagenet}
\subsection{Main Results}\label{sec:imagenet}

The results of different CIL methods on both the ImageNet100 and ImageNet1000 are shown in \cref{tab:imagenet}.
% The results of different methods on the ImageNet100 are shown in \cref{tab:imagenet} and \cref{fig:imagenet}.
The proposed MAFDRC achieves the same level of performance as the strong competitors, i.e., FOSTER~\cite{foster} and DER~\cite{der}, and becomes one of the state-of-the-art (SOTA) methods.
Our method consistently surpasses the FOSTER and its uncompressed variant, FOSTER B4, with clear margins.
For example, MAFDRC is better than FOSTER B4 with {$3.12\%$ and $1.68\%$} improvements in the averaged accuracy under the B0 10 steps and the B50 10 steps of ImageNet100, respectively. On the large-scale ImageNet1000, such improvement is still more than $1\%$.
Similar results are obtained by MAFDRC and the DER w/o P (DER without pruning). 
Note that DER w/o P is a strong CIL method that keeps expanding its feature extractors and ends up with a much larger model size than ours with the fixed numbers of parameters and feature dimensions across tasks.
% Note that DER w/o P is a strong CIL method with a much larger model size than our method. 
However, MAFDRC still achieves better results than DER w/o P in multiple settings, including the challenging ImangeNet1000, with nontrivial improvement (usually more than $1\%$).
Detailed comparisons among different methods along the incremental learning procedure are illustrated in \cref{fig:imagenet}.

The CIL results on CIFAR100 are shown in \cref{tab:cifar} and \cref{fig:cifar}.
DER achieves the best performance at the price of a much larger model size than the proposed method, i.e., the model parameters in DER are rough $k$ times of those in MAFDRC under $k$ steps.
Our method achieves the second-best results
% in most cases 
and is better than other SOTA methods with similar model sizes, e.g., FOSTER.
The proposed method is also validated on a new CIL setting, LT-CIL~\cite{liu2022long}, with other SOTA results reported in \cref{tab:LT-CIL}.
% Comparisons among different SOTA methods can be found in \cref{tab:LT-CIL}.
Our MAFDRC consistently achieves the new SOTA results on both datasets and settings.
More importantly, the proposed method boosts its baseline, the MAF pipeline, with more than {$10 \%$} in most cases.
% on all criteria. 
It suggests the effectiveness of the proposed dynamic residual classifier (DRC) in handling the data imbalance issue.

% \cref{tab:LT-CIL} shows the results on LT-CIL setting that is a more challenging scenario. The MAF pipeline with DRC achieves significantly better accuracy compared to other methods for Ordered LT-CIL and Shuffled LT-CIL. It suggests that the DRC is as robust as not only in the conventional CIL scenario, but in the more challenging LT-CIL scenario.

%------------------------------------------------------------------------
% \subsection{Ablation Studies}\label{sec:ablation}
\subsection{Detailed Analysis}\label{sec:ablation}

% {\color{red}{
The results of CIFAR100 {\bf B0} 10 steps and ImageNet100 {\bf B0} 10 steps are reported by default.
% }}
% More analysis is in the Supplementary Material.
More results can be found in the Supplementary Material.
% In this section, the experimental results are based on the CIFAR100/ImageNet100 B0 10 steps. More results on CIFAR100/ImageNet100 B0 5,20 steps are in the supplementary.
% we first evaluate the contribution of each component for our method. Then, we compare our proposed model adaptation and fusion pipeline with other pipelines. The experiment results of MEC are those after model compression. 
% CIL setting is by default, 

% \usepackage{multirow}
\begin{table}[t]
\centering
% one run on seed 1993
% \begin{tabular}{ccccc}
% \hline
% \multicolumn{3}{c}{{Components}}   & \multirow{2}{*}{{Avg}} & \multirow{2}{*}{{Last}} \\
% {MAF} & {DRC} & {LA} &                               &                                \\ \hline
%              &              &             & 65.59                  & 50.56                   \\
% \checkmark            &              &             & 70.53                  & 55.71                    \\
% \checkmark            & \checkmark            &             & 72.80                         & 60.92                          \\
% \checkmark            & \checkmark            & \checkmark           & 73.97                         & 62.04                          \\ \hline

% \end{tabular}
\begin{tabular}{ccc|cccc}
\hline
\multicolumn{3}{c|}{Components} & \multicolumn{2}{c}{CIFAR100} & \multicolumn{2}{c}{ImageNet100} \\ \hline
MAF       & DRC       & LA      & Avg           & Last         & Avg            & Last           \\ \hline
         &          &        & 65.59         & 50.56        & 68.95          & 54.18          \\
\checkmark          &          &        & 69.61         & 54.84        & 72.01          & 58.76          \\
\checkmark         & \checkmark          &        & 72.80         & 60.92        & 78.65          & 69.88          \\
\checkmark         & \checkmark          & \checkmark        & 73.97         & 62.04        & 79.66          & 70.41          \\ \hline
\end{tabular}
\caption{
% {\color{red}{
Contributions of different components in MAFDRC.
% }}
% {\color{red}{The contributions of different model components on the proposed MAFDRC.}}
% Ablative Study on the proposed MAFDRC.
% The contribution of each component. 
% MAF means the model adaptation and fusion pipeline. DRC means the dynamic residual classifier. LA means the logits adjustment~\cite{menon2020long}.
}
\label{tab:component}
\end{table}

\begin{table}[t]
\centering
\begin{tabular}{c|cccc}
\hline
\multirow{2}{*}{Methods} & \multicolumn{2}{c}{CIFAR100}   & \multicolumn{2}{c}{ImageNet100}                    \\ \cline{2-5} 
                         & Avg & Last & Avg & Last \\ \hline
MAF                    & 69.61                       & 54.84    & 72.01  & 58.76   \\
MAFRC                    & 74.07                       & 61.91    & 79.26  & 70.54   \\
MAFDRC                   & 73.97                       & 62.04    & 79.66  & 70.41  \\ \hline
\end{tabular}
\caption{The results of MAF with residual classifier (RC) and dynamic residual classifier (DRC).}
\label{tab:RC}
\end{table}

% {\color{red}{
\noindent {\bf Contributions of different components} \quad
% { \bf Effect of each component } \quad \cref{tab:component} 
The results are shown in \cref{tab:component}.
% }}
The MAF pipeline achieves clearly better results than the finetuning baseline since such a pipeline is a CIL method itself, as discussed in \cref{sec:relat}.
% Substantial improvements are brought about by the proposed DRC.
The proposed DRC brings substantial improvements. For example, DRC enhances MAF results with $6.64\%$ average accuracy and $11.12\%$ last accuracy on ImageNet100.
% The proposed DRC is compatible with the MAF pipeline. Their combination is denoted as MAF+DRC.
By replacing the conventional classification loss with the adjusted loss, logit adjustment (LA)~\cite{menon2020long}, provides positive effects and results in the proposed method, MAFDRC.
% \cref{tab:component} demonstrates the results of our ablative experiments on CIFAR100 and ImageNet100 with 10 steps. The baseline method is only use cross-entropy loss with $D_t$ and $M_t$. We can see that the average accuracy is improved significantly from 65.59\% to 72.80\% by the combination with DRC and MAF. We also see that the performance of the model is further improved with 1.17\% gain using logits adjustment.
% % difference ````, 还需解释一下为啥MAFDRC会更好一点

\noindent {\bf Branch Layer Merging} \quad 
% { \bf The effect of branch layer merging} \quad 
As shown in \cref{tab:RC}, MAFRC clearly improves the baseline. Therefore, the effectiveness of the branch layer architecture and residual fusion mechanism in RC is demonstrated.
% achieves similar results to MAFDRC. Therefore, the branch layer architecture and residual fusion mechanism in RC are effective under the CIL settings.
However, with the increasing task-specific branch layers, MAFRC suffers from the problem of growing model size.
DRC handles this issue with branch layer merging, resulting in a more efficient model, MAFDRC.
Moreover, MAFDRC is as effective as MAFRC, as suggested by their performance in \cref{tab:RC}.

\begin{table}[t]
% \centering
% \begin{center}
\scalebox{0.8}{
\begin{tabular}{c|llll}
\hline
\multirow{2}{*}{Model} & \multicolumn{2}{l}{\ \ \ \ \ \ \ \ \ \ CIFAR100} & \multicolumn{2}{l}{\ \ \ \ \ \ \ ImageNet100} \\ \cline{2-5} 
                       & Avg           & Last         & Avg            & Last           \\ \hline
MAF                    & 69.61         & 54.84        & 72.01               & 58.76               \\
+DRC                   & 72.80{\color[HTML]{FE0000} ($\uparrow$3.19)}         & 60.92{\color[HTML]{FE0000} ($\uparrow$6.08)}        & 78.65{\color[HTML]{FE0000} ($\uparrow$6.64)}               & 69.88{\color[HTML]{FE0000} ($\uparrow$11.12)}              \\ \hline
MEC                    & 69.02         & 52.86        & 70.08               &    55.06            \\
+DRC                   & 70.25{\color[HTML]{FE0000} ($\uparrow$1.23)}         & 54.76{\color[HTML]{FE0000} ($\uparrow$1.90)}        & 71.41{\color[HTML]{FE0000} ($\uparrow$1.33)}               &    58.54{\color[HTML]{FE0000} ($\uparrow$3.48)}            \\ \hline
MDT                    & 67.69         & 52.56        & 69.40               & 54.68               \\
+DRC                   & 70.80{\color[HTML]{FE0000} ($\uparrow$3.11)}         & 58.28{\color[HTML]{FE0000} ($\uparrow$5.72)}        & 71.31{\color[HTML]{FE0000} ($\uparrow$1.91)}               & 59.34{\color[HTML]{FE0000} ($\uparrow$4.66)}               \\ \hline
\end{tabular}}
% \end{center}
\caption{The results of different pipelines with DRC.}
\label{tab:pipeline all data}
\end{table}

\noindent {\bf CIL Pipelines with DRC} \quad
% { \bf The results of pipelines with DRC} \quad
As shown in \cref{tab:pipeline all data},
% all three CIL pipelines benefit the most from integrating DRC
DRC is compatible with all three pipelines and clearly boosts their performance. MAF benefits the most.

\begin{table}[t]
\centering
\begin{tabular}{c|cccc}
\hline
\multirow{2}{*}{Methods} & \multicolumn{2}{c}{CIFAR100} & \multicolumn{2}{c}{ImageNet100}                     \\ \cline{2-5} 
                         & Avg & Last & Avg & Last \\ \hline
MAF                    & 69.61                       & 54.84    & 72.01  & 58.76   \\
MAF+BFT~\cite{kang2019decoupling}                    & 69.75                       & 55.19    & 73.08   & 59.96  \\
MAF+WA~\cite{wa}                    & 70.14                       & 56.23    & 73.94  & 62.14  \\
% MAFRC                    & 75.12                       & 73.89    & 71.23    \\
MAF+DRC                   & \textbf{72.80}                       & \textbf{60.92}    & \textbf{78.65}  & \textbf{69.88}   \\ \hline
\end{tabular}
\caption{The results of different imbalanced methods.}
\label{tab:RC others}
\end{table}

\noindent {\bf Data Imbalanced Methods} \quad 
% { \bf Compared with other imbalanced methods} \quad 
The balanced fine-tuning (BFT)~\cite{kang2019decoupling} and  the weight aligning (WA)~\cite{wa} are alternatives to DRC for handling the data imbalance in CIL. As shown in \cref{tab:RC others}, both BFT and WA bring some improvements to the MAF pipeline but are inferior to our DRC.
% \cref{tab:RC others} shows the effectiveness of DRC both on CIFAR100 and ImageNet100. BFT means the balanced fine-tuned method~\cite{kang2019decoupling}. WA means the weight aligning method~\cite{wa}.

% \begin{table}[t]
% \centering
% \begin{tabular}{c|ccc|c}
% \hline
% \multirow{2}{*}{Methods} & \multicolumn{3}{c|}{CIFAR100} & ImageNet100                     \\ \cline{2-5} 
%                          & 5 steps & 10 steps & 20 steps & 10 steps \\ \hline
% MAF+BFT                    & 71.24                       & 69.75    & 67.01   & 73.08  \\
% MAF+WA                    & 71.39                       & 70.14    & 66.90  & 73.94  \\
% % MAFRC                    & 75.12                       & 73.89    & 71.23    \\
% MAF+DRC                   & \textbf{74.87}                       & \textbf{73.97}    & \textbf{71.75}  & \textbf{79.66}   \\ \hline
% \end{tabular}
% \caption{DRC\&RC,BFT,WA.}
% \label{tab:RC others}
% \end{table}

\begin{figure}[t]
\centering
  \begin{subfigure}{0.48\linewidth}
    % \fbox{\rule{0pt}{2in} \rule{.9\linewidth}{0pt}}
    \includegraphics[width=1\linewidth]{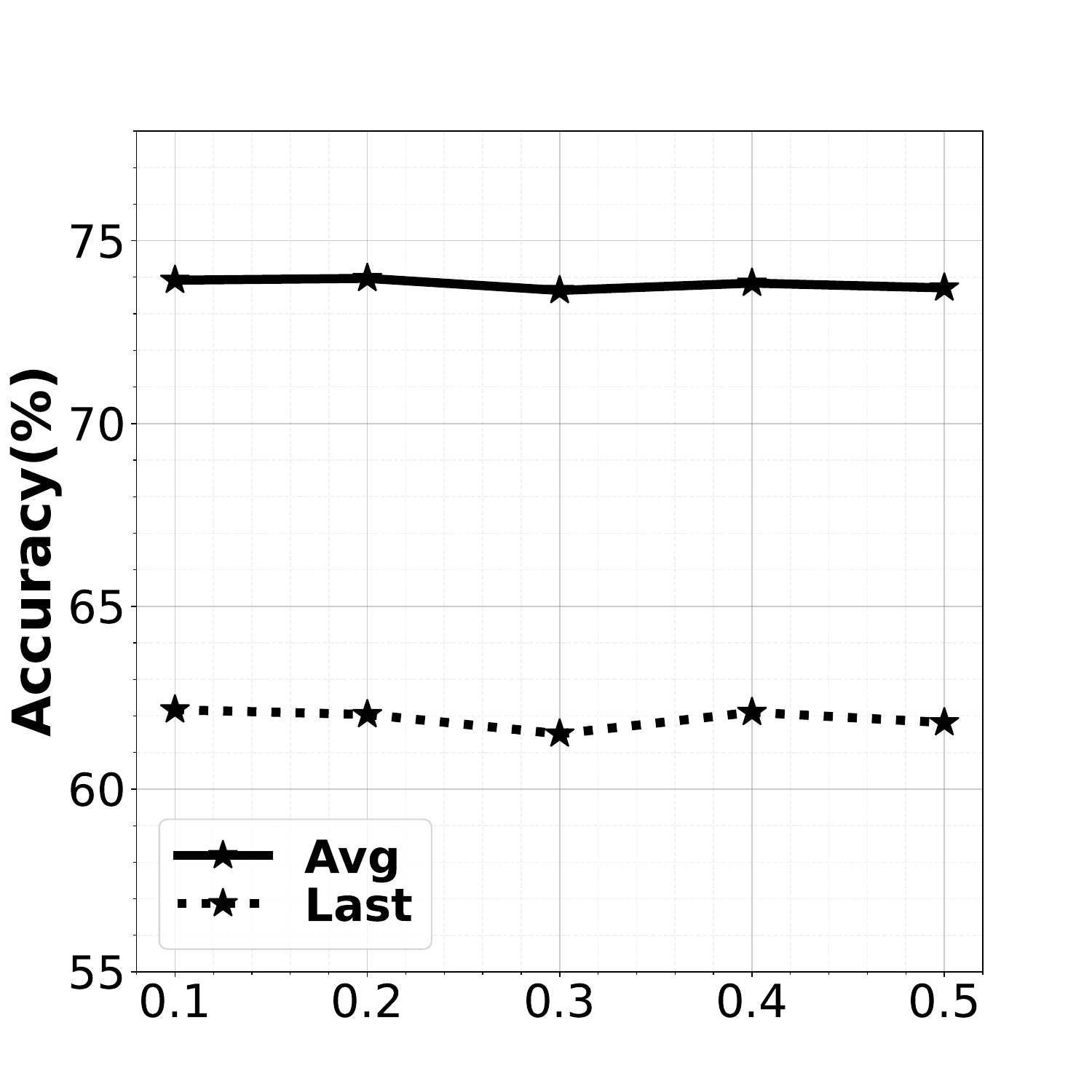}
    \caption{$\alpha$}
    \label{fig:alpha}
  \end{subfigure}
  \hfill
  \begin{subfigure}{0.48\linewidth}
    % \fbox{\rule{0pt}{2in} \rule{.9\linewidth}{0pt}}
    \includegraphics[width=1\linewidth]{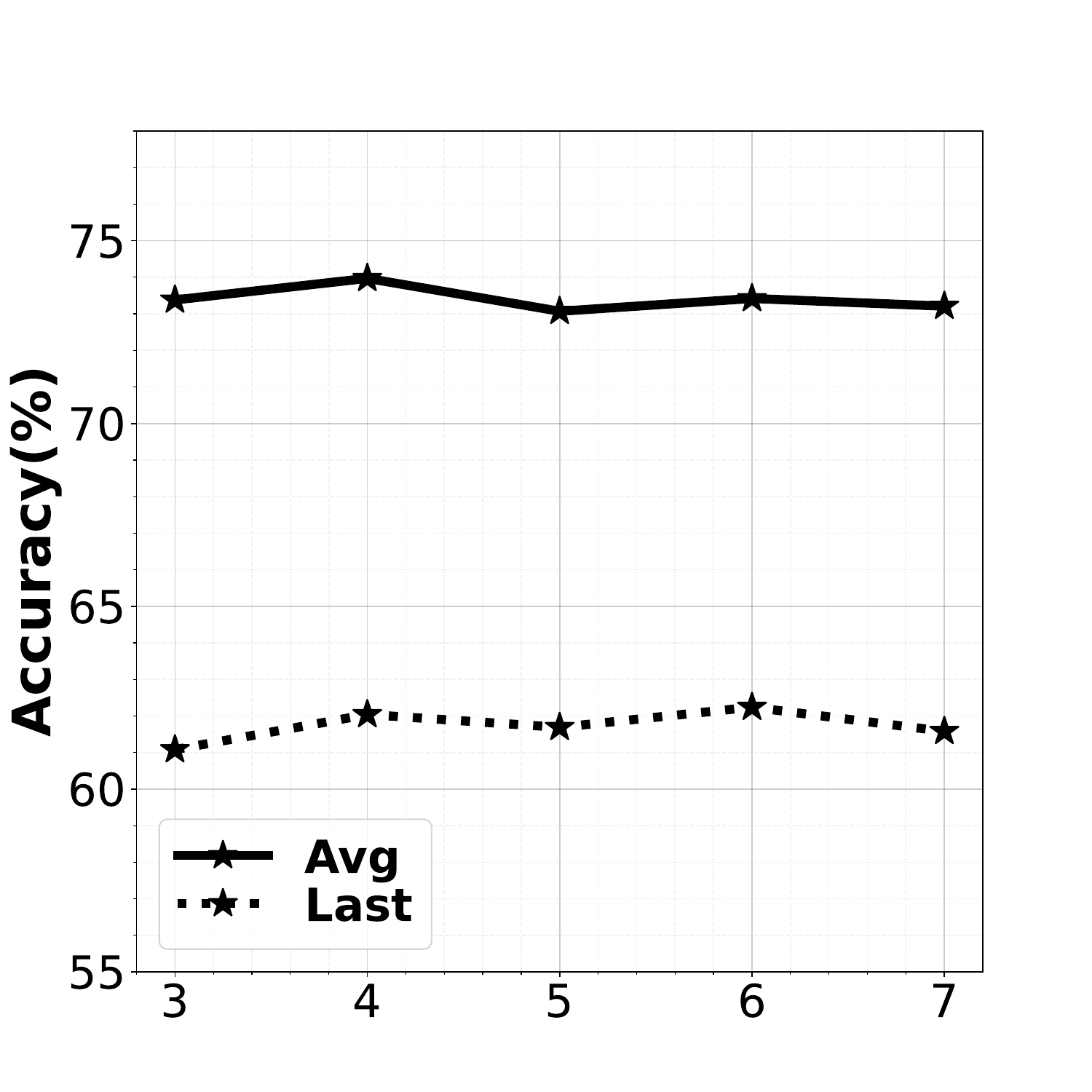}
    \caption{$\beta$}
    \label{fig:beta}
  \end{subfigure}
  % \hfill
  % \begin{subfigure}{0.32\linewidth}
  %   % \fbox{\rule{0pt}{2in} \rule{.9\linewidth}{0pt}}
  %   \includegraphics[width=1\linewidth]{figs/cifar100_results_parameteratro4.pdf}
  %   \caption{tro.}
  %   \label{fig:tro}
  % \end{subfigure}
  \caption{
  Impacts of $\alpha$s and $\beta$s.
  % on our method. 
  CIFAR100 B0 10 steps is used.
  % The Sensitive study of hyper-parameters.
  % Sensitive study of hyper-parameters.
  }
  % The results are "Avg" top-1 accuracy (\%) (above) and "Last" top-1 accuracy (below) for different parameters.}
  \label{fig:parameter}
\end{figure}

\noindent {\bf Hyper-parameters} \quad
% \noindent {\bf  Sensitivity of Hyper-parameter} \quad
% { \bf Sensitive study of hyper-parameters} \quad
The memory size of the B0 protocol~\cite{icarl} is reduced to 20 exemplars per class. The results are reported in \cref{tab:20exemplar}. Our MAFDRC still achieves the SOTA level performance.
The performance of our methods with different hyper-parameter values 
% different balancing hyper-parameters $\alpha$ and $\beta$ in 
of \cref{eq:sum} are shown in \cref{fig:parameter}. Our method is not sensitive to such changes.
% on a wide range.

% \cref{fig:parameter} verifies the robustness of MAFDRC on CIFAR100 10steps with different hyper-parameters $\alpha$ and $\beta$. We test $\alpha=0.1, 0.2, 0.3, 0.4, 0.5$ and $\beta=3, 4, 5, 6, 7$, respectively. We can see that the performance changes are minimal under different $\alpha$s and $\beta$s.
% % \noindent {\bf Rehearsal Memory Sizes} \quad 
% % { \bf Effect of number of exemplars}\quad 
% In \cref{tab:20exemplar}, the number of exemplars are changed to 20 exemplars per class and no more than 2,000 exemplars. This indicates that pipeline with DRC is effective and robust; it can overcome forgetting even with more challenging exemplars reserved.

% \usepackage{multirow}
\begin{table}[t]
\centering
% \footnotesize
% \scalebox{0.7}
{
\begin{tabular}{c|cccc}
\hline
\multirow{2}{*}{Methods} & \multicolumn{2}{c}{CIFAR100} & \multicolumn{2}{c}{ImageNet100} \\ \cline{2-5} 
                         & Avg  & Last  & Avg  & Last     \\ \hline
DER w/o P                & \color[HTML]{FE0000} \textbf{71.99}       & \color[HTML]{FE0000} \textbf{62.66}        & \color[HTML]{3166FF} \textbf{76.49}
       & \color[HTML]{3166FF} \textbf{69.42}           \\
FOSTER B4                & 71.16       & 59.40        & 75.63        & 64.90           \\
FOSTER                   & 69.79       & 58.33        & 74.58        & 64.72           \\ 
MAFDRC                   & \color[HTML]{3166FF} \textbf{71.24}       & \color[HTML]{3166FF} \textbf{59.73}        & \color[HTML]{FE0000} \textbf{77.88}        & \color[HTML]{FE0000} \textbf{69.64}           \\ \hline
\end{tabular}}
% \vspace{-0.1cm}
\caption{CIL results with reserving 20 exemplars per class.}
\label{tab:20exemplar}
\end{table}

\section{Conclusion}
In this paper, the dynamic nature of the data imbalance in the widely used CIL rehearsal strategy is shown.
We aim to handle this challenging scenario with a novel dynamic residual classifier (DRC). This is complementary to the adjusted losses and data re-samplings used by many CIL methods based on the static viewpoint.
The proposed DRC adopts the branch layer architecture and the residual fusion mechanism of a recent advance residual classifier (RC) and handles the model-growing problem with the branch layer merging.
As a generalizable method, DRC substantially improves the performance of different CIL pipelines and achieves SOTA performance under both the CIL and LT-CIL settings.

\noindent {\bf Acknowledgement} \quad
This research is supported partly by the National Science Foundation for Young Scientists of China (No. 62106289), partly by National Natural Science Foundation of China (No. 62232008), partly by Guangdong HUST Industrial Technology Research Institute, Guangdong Provincial Key Laboratory of Manufacturing Equipment Digitization (2020B1212060014).

{\small
\bibliographystyle{ieee_fullname}
\bibliography{egbib}
}

\end{document}

%%%%%%%%% TITLE
\title{Supplementary Material of\\
Dynamic Residual Classifier for Class Incremental Learning}

% \and

% \maketitle
\ificcvfinal\thispagestyle{empty}\fi

% \section{Additional Main Results}\label{supp:add_results}
\section{Additional Main Results}\label{supp_sec:add_results}
% Fig.~\ref{main-fig:frame}
% Fig.~\ref{fig:frame}

% The results in Tab.~\ref{tab:imagenet without da} and Tab.~\ref{tab:cifar-without da} are reproduced without the AutoAugment~\cite{cubuk2019autoaugment} as in FOSTER~\cite{foster}.

% In this section, the CIL results without the AutoAugment~\cite{cubuk2019autoaugment} as in FOSTER~\cite{foster} are reported.

In the main text, different CIL methods are trained with AutoAugment~\cite{cubuk2019autoaugment}. Their results are reported in Tab.~\ref{tab:imagenet} and Tab.~\ref{tab:cifar}.
In this section, we reproduce the corresponding results without such data augmentation, as shown in Tab.~\ref{tab:imagenet without da} and Tab.~\ref{tab:cifar-without da}, respectively.
The proposed MAFDRC still achieves the same level of performance as the state-of-the-art methods, e.g., FOSTER~\cite{foster} and DER~\cite{der}. In particular, under the CIL setting of ImageNet1000 10 steps,  our method achieves similar results compared to DER w/o P but with a much smaller model size.

\section{Additional Detailed Analysis Results}\label{supp_sec:add detail}
% tro equation &analysis

Detailed analysis is conducted mainly on the CIFAR100 and ImageNet100 B0 10 steps in Sec.~\ref{sec:ablation}.
More results on CIFAR100/ImageNet100 B0 5, 20 steps are presented here (with average accuracy only).

% In the Sec. \ref{sec:ablation} of the main text, we provide the detailed analysis on CIFAR100/ImageNet100 B0 10 steps. Here we provide more results on CIFAR100/ImageNet100 B0 5, 20 steps with AutoAugment.

% \subsection{Ablation Study}\label{supp_sec:each component}

% \noindent {\bf Ablation Study} \quad
% Tab.~\ref{supp_tab:component} shows the contributions of different components in the proposed method.
% % on CIFAR100/ImageNet100 5,20 steps. 
% Clear improvements can be observed by introducing the proposed DRC.
% % The proposed DRC improves the average accuracy significantly with a similar results in the main text.  

% {\color{red}{
\noindent {\bf Contributions of Different Components} \quad
As shown in Tab.~\ref{supp_tab:component}, clear improvements can be observed by introducing the proposed DRC.
% }}

% \usepackage{multirow}
\begin{table}[ht]
\centering
\scalebox{0.9}{
% one run on seed 1993
% \begin{tabular}{ccccc}
% \hline
% \multicolumn{3}{c}{{Components}}   & \multirow{2}{*}{{Avg}} & \multirow{2}{*}{{Last}} \\
% {MAF} & {DRC} & {LA} &                               &                                \\ \hline
%              &              &             & 65.59                  & 50.56                   \\
% \checkmark            &              &             & 70.53                  & 55.71                    \\
% \checkmark            & \checkmark            &             & 72.80                         & 60.92                          \\
% \checkmark            & \checkmark            & \checkmark           & 73.97                         & 62.04                          \\ \hline

% \end{tabular}
\begin{tabular}{ccc|cccc}
\hline
\multicolumn{3}{c|}{Components} & \multicolumn{2}{c}{CIFAR100} & \multicolumn{2}{c}{ImageNet100} \\ \hline
MAF       & DRC       & LA      & 5 steps           & 20 steps         & 5 steps            & 20 steps           \\ \hline
         &          &        & 65.82         & 64.66        & 70.62          & 66.53          \\
\checkmark          &          &        & 71.24         & 67.01        & 77.49          & 68.50          \\
\checkmark         & \checkmark          &        & 74.43         & 71.50        & 82.16          & 73.67          \\
\checkmark         & \checkmark          & \checkmark        & 74.87         & 71.75        & 82.22          & 75.21          \\ \hline
\end{tabular}}
\caption{
% {\color{red}{
The contributions of different model components on the proposed MAFDRC.
% }}
% Ablation Study on the proposed MAFDRC. 
Corresponding to Tab. \ref{tab:component} in the main text.}
\label{supp_tab:component}
\end{table}

% \subsection{Branch Layer Merging}\label{supp_sec:blm}
\noindent {\bf Branch Layer Merging} \quad
Tab. \ref{supp_tab:RC} further demonstrates the effectiveness of the residual classifier architecture and the proposed branch layer merging.
% shows the effectiveness of the branch layer merging. On different experimental settings, the simple branch layer merging also achieves the similar results.

\begin{table}[ht]
\centering
\begin{tabular}{c|cccc}
\hline
\multirow{2}{*}{Methods} & \multicolumn{2}{c}{CIFAR100}   & \multicolumn{2}{c}{ImageNet100}                    \\ \cline{2-5} 
                         & 5 steps & 20 steps & 5 steps & 20 steps \\ \hline
MAF                      & 71.24                       & 67.01    & 77.49  & 68.50
\\
MAFRC                    & 75.12                       & 71.23    & 82.51  & 75.11   \\
MAFDRC                   & 74.87                       & 71.75    & 82.22  & 75.21  \\ \hline
\end{tabular}
\caption{The results of MAF with residual classifier and dynamic residual classifier. Corresponding to Tab. \ref{tab:RC} in the main text.}
\label{supp_tab:RC}
\end{table}

% \subsection{CIL Pipelines with DRC}\label{supp_sec:pipeline DRC}
\noindent {\bf CIL Pipelines with DRC} \quad
As shown in Tab. \ref{supp_tab:pipeline all data}, DRC consistently improves all pipelines under the new settings.
% is also compatible with all three pipelines on the CIFAR100/ImageNet100 B0 5, 20 steps.

\begin{table}[t]
\centering
% \begin{center}
\scalebox{0.8}{
\begin{tabular}{cllll}
\hline
\multirow{2}{*}{Model} & \multicolumn{2}{l}{\ \ \ \ \ \ \ \ \ \ CIFAR100} & \multicolumn{2}{l}{\ \ \ \ \ \ \ ImageNet100} \\ \cline{2-5} 
                       & 5 steps           & 20 steps         & 5 steps            & 20 steps           \\ \hline
MAF                    & 71.24         & 67.01        & 77.49               & 68.50             \\
+DRC                   & 74.43{\color[HTML]{FE0000} ($\uparrow$3.19)}         & 71.50{\color[HTML]{FE0000} ($\uparrow$4.49)}        & 82.16{\color[HTML]{FE0000} ($\uparrow$4.67)}               & 73.67{\color[HTML]{FE0000} ($\uparrow$5.17)}             \\ \hline
MEC                   & 69.02         & 67.52        & 71.49               & 67.61             \\
+DRC                   & 70.64{\color[HTML]{FE0000} ($\uparrow$1.62)}         & 69.29{\color[HTML]{FE0000} ($\uparrow$1.77)}        & 73.60{\color[HTML]{FE0000} ($\uparrow$2.11)}               & 68.40{\color[HTML]{FE0000} ($\uparrow$0.79)}             \\\hline
MDT                    & 68.24         & 67.01        & 73.07               & 67.12             \\
+DRC                   & 70.08{\color[HTML]{FE0000} ($\uparrow$1.84)}         & 68.51{\color[HTML]{FE0000} ($\uparrow$1.50)}        & 73.18{\color[HTML]{FE0000} ($\uparrow$0.11)}               & 68.39{\color[HTML]{FE0000} ($\uparrow$1.27)}             \\\hline
\end{tabular}}
% \end{center}
\caption{The results of different pipelines with DRC. Corresponding to Tab. \ref{tab:pipeline all data} in the main text.}
\label{supp_tab:pipeline all data}
\end{table}

% \subsection{Data Imbalanced Methods}\label{supp_sec:imba methods}
\noindent {\bf Data Imbalanced Methods} \quad
As shown in Tab. \ref{supp_tab:RC others}, DRC is superior to its counterparts, BFT and WA, in handling the data imbalance in CIL.
% gets the more improvements compared to both BFT and WA. This also illustrates the superiority of our method.

\begin{table}[t]
\centering
\begin{tabular}{c|cccc}
\hline
\multirow{2}{*}{Methods} & \multicolumn{2}{c}{CIFAR100} & \multicolumn{2}{c}{ImageNet100}                     \\ \cline{2-5} 
                         & 5 steps & 20 steps & 5 steps & 20 steps \\ \hline
MAF                      & 71.24                       & 67.01    & 77.49  & 68.50
\\
MAF+BFT                    & 71.36                       & 67.58    & 78.21   & 68.76  \\
MAF+WA                    & 71.39                       & 67.70    & 79.62   & 68.71  \\
% MAFRC                    & 75.12                       & 73.89    & 71.23    \\
MAF+DRC                   & \textbf{74.43}                 & \textbf{71.50}    & \textbf{82.16}   & \textbf{73.67}  \\ \hline
\end{tabular}
\caption{The results of different imbalanced methods. Corresponding to Tab. \ref{tab:RC others} in the main text.}
\label{supp_tab:RC others}
\end{table}

% \subsection{Hyper-parameters}\label{supp_sec:hp}
\noindent {\bf Hyper-parameters} \quad
Tab.~\ref{supp_tab:20exemplar} shows more results under the memory size of 20 exemplars per class.
The results under the memory size of 1,000 exemplars are shown in Tab.~\ref{supp_tab:1000exemplar}.
% Tab.~\ref{supp_tab:20exemplar} and Tab.~\ref{supp_tab:1000exemplar} show the results of the dynamic residual classifier with MAF pipeline on different exemplars, 20 exemplars per class (no more than 2,000 exemplars) and 1,000 exemplars in total.
% % This indicates that the pipeline with DRC is effective and robust; it can overcome forgetting not only even with fewer exemplars but a fixed exemplars per class.

\begin{table}[t]
\centering
% \footnotesize
% \scalebox{0.7}
{
\begin{tabular}{c|cccc}
\hline
\multirow{2}{*}{Methods} & \multicolumn{2}{c}{CIFAR100} & \multicolumn{2}{c}{ImageNet100} \\ \cline{2-5} 
                         & 5 steps  & 20 steps  & 5 steps  & 20 steps     \\ \hline
DER w/o P                & \color[HTML]{3166FF} \textbf{74.10}       & \color[HTML]{FE0000} \textbf{67.06}        & \color[HTML]{3166FF} \textbf{79.85}         & \color[HTML]{FE0000} \textbf{72.97}           \\
FOSTER B4                & \color[HTML]{FE0000} \textbf{74.55}       &  66.65       & 78.64         & \color[HTML]{3166FF} \textbf{71.31}           \\
FOSTER                    & 72.34       & 66.06        & 76.62         & 70.45           \\
MAFDRC                   & 73.19       & \color[HTML]{3166FF} \textbf{66.91}        & \color[HTML]{FE0000} \textbf{81.31}         & 70.07           \\ \hline
\end{tabular}}
% \vspace{-0.1cm}
\caption{CIL results with reserving 20 exemplars per class. Corresponding to Tab. \ref{tab:20exemplar} in the main text.}
\label{supp_tab:20exemplar}
\end{table}

% \vspace{-0.2cm}
\begin{table}[t]
\centering
% \footnotesize
\scalebox{0.7}
{
\begin{tabular}{c|cccccc}
\hline
\multirow{2}{*}{Methods} & \multicolumn{3}{c}{CIFAR100} & \multicolumn{3}{c}{ImageNet100} \\ \cline{2-7} 
                         & 5 steps & 10 steps  & 20 steps  & 5 steps & 10 steps & 20 steps \\ \hline
DER w/o P                 & \color[HTML]{FE0000} \textbf{74.37}  & \color[HTML]{FE0000} \textbf{73.07}      & \color[HTML]{FE0000} \textbf{72.57}        & \color[HTML]{3166FF} \textbf{79.67}   & \color[HTML]{3166FF} \textbf{76.90}      & \color[HTML]{FE0000} \textbf{75.54}           \\
FOSTER B4                 & 73.04  & 71.16     & 67.86        & 75.58   & 72.99      & 71.41           \\
FOSTER                    & 70.34   & 69.63    & 66.92        & 72.88   & 71.55      & 70.48           \\
MAFDRC                    & \color[HTML]{3166FF} \textbf{73.46}   & \color[HTML]{3166FF} \textbf{71.62}    & \color[HTML]{3166FF} \textbf{67.91}        & \color[HTML]{FE0000} \textbf{81.20}    & \color[HTML]{FE0000} \textbf{78.39}     & \color[HTML]{3166FF} \textbf{72.37}           \\ \hline
\end{tabular}}
% \vspace{-0.1cm}
\caption{CIL results with a memory size of 1,000 exemplars.}
\label{supp_tab:1000exemplar}
\end{table}

% \begin{table}[t]
% \centering
% \scalebox{0.9}{
% \begin{tabular}{c|cccccc}
% \hline
% \multirow{2}{*}{Steps} & \multicolumn{6}{c}{$tro$}   \\ 
% % \multirow{2}{*}{Steps} & \multicolumn{6}{c}{CIFAR100 B0}   \\ \cline{2-7} 
%                        & 1.0 & 1.2 & 1.4 & 1.6 & 1.8 & 2.0 \\ \hline
% 5   & 74.69 & 74.87 & 74.28 & 73.80 & 73.78 & 72.26 \\
% 10  & 73.76 & 73.97 & 73.30 & 73.40 & 72.56 & 72.04 \\
% 20  & 71.04 & 71.75 & 71.99 & 71.75 & 71.76 & 71.00 \\ \hline
% \end{tabular}}
% \caption{Impacts of different $tro$s on model performance. CIFAR100 B0 20 steps results are reported.
% % with average accuracy reported.
% }
% \label{supp_tab:tro}
% \end{table}
% % tro的位置，B0 

\begin{table}[t]
\centering
\scalebox{0.9}{
\begin{tabular}{ccccccc}
\hline
tro & 1.0 & 1.2 & 1.4 & 1.6 & 1.8 & 2.0 \\
 % 74.69 & 74.87 & 74.28 & 73.80 & 73.78 & 72.26 \\
 % 73.76 & 73.97 & 73.30 & 73.40 & 72.56 & 72.04 \\
Avg & 71.04 & 71.75 & 71.99 & 71.75 & 71.76 & 71.00 \\ \hline
\end{tabular}}
\caption{Impacts of different $tro$s on model performance. CIFAR100 B0 20 steps results are reported.
% with average accuracy reported.
}
\label{supp_tab:tro}
\end{table}

\begin{table*}[t]
\centering
% \scalebox{1.0}
% {
\begin{tabular}{c|cccccc|cccc|cc}
\hline
\multirow{3}{*}{Methods} & \multicolumn{6}{c|}{ImageNet100 B0}                                                        & \multicolumn{4}{c|}{ImageNet100 B50}                        & \multicolumn{2}{c}{ImageNet1000} \\ \cline{2-13} 
                         & \multicolumn{2}{c}{5 steps} & \multicolumn{2}{c}{10 steps} & \multicolumn{2}{c|}{20 steps} & \multicolumn{2}{c}{5 steps} & \multicolumn{2}{c|}{10 steps} & \multicolumn{2}{c}{10 steps}     \\ \cline{2-13} 
                         & Avg          & Last         & Avg          & Last          & Avg           & Last          & Avg          & Last         & Avg           & Last          & Avg            & Last            \\ \hline
iCaRL~\cite{icarl}    & 71.25                                 & 60.02                                  & 65.82                                 & 50.86                                  & 61.07                                 & 44.66                                      & 58.90   & 49.34  & 48.59  & 41.70  & 53.32  & 33.96                          \\
BiC~\cite{wu2019large}       & 71.24                                 & 60.98                                & 64.93                                 & 45.98                              & 56.18                                 & 32.32       & 61.65   & 44.98  & 53.71  & 37.38  & -  & -                           \\
WA~\cite{wa}        & 73.89                                 & 64.06                                 & 68.00                                 & 54.48                                & 61.96                                 & 46.02       & 62.78   & 54.24  & 52.84  & 45.70  & -  & -                           \\
PODNet~\cite{podnet}    & 72.53                                 & 58.68                                  & 62.85                                 & 44.84                                 & 54.88                                 & 36.86       & 75.33   & 66.58  & 72.91  & 62.90  & -  & -                            \\ 
DER w/o p~\cite{der} & {\color[HTML]{3166FF} \textbf{77.57}}                                 & {\color[HTML]{3166FF} \textbf{71.28}}                                                               & {\color[HTML]{FE0000} \textbf{75.49}}                                 & {\color[HTML]{FE0000} \textbf{66.34}}                                                   & {\color[HTML]{FE0000} \textbf{72.87}}                                 & {\color[HTML]{FE0000} \textbf{64.92}}       & {\color[HTML]{3166FF} \textbf{77.36}}   & {\color[HTML]{3166FF} \textbf{70.82}}  & {\color[HTML]{FE0000} \textbf{75.59}}  & {\color[HTML]{FE0000} \textbf{68.94}}  & {\color[HTML]{3166FF} \textbf{66.74}}  & {\color[HTML]{FE0000} \textbf{57.84}}     \\
% FOSTER B4~\cite{foster} & 75.81 & 69.66         & 71.37 & 62.58                 & 66.42 & {\color[HTML]{3166FF} \textbf{54.18}}           & -   & -  & 73.75  & 63.98  & {\color[HTML]{38FFF8} \textbf{64.15}}  & {\color[HTML]{38FFF8} \textbf{45.03}}      \\
% FOSTER~\cite{foster}    & 74.90                                 & 68.54                                 & 71.07                                 & 63.14                           & 66.48                                 & 55.34       & -   & -  & 73.86  & 64.50  & {\color[HTML]{38FFF8} \textbf{63.14}}  & {\color[HTML]{38FFF8} \textbf{44.90}}     \\ \hline
FOSTER B4~\cite{foster} & 75.81 & 69.66         & 71.37 & 62.58                 & 66.42 & {\color[HTML]{3166FF} \textbf{54.18}}           & 77.15   & 70.06  & 73.75  & 63.98  & 64.15  & 45.03      \\
FOSTER~\cite{foster}    & 74.90                                 & 68.54                                 & 71.07                                 & 63.14                           & 66.48                                 & 55.34       & 76.89   & 70.92  & 73.86  & 64.50  & 63.14  & 44.90     \\ \hline
MAFDRC    & {\color[HTML]{FE0000} \textbf{78.48}} & {\color[HTML]{FE0000} \textbf{71.28}} & {\color[HTML]{3166FF} \textbf{74.31}} & {\color[HTML]{3166FF} \textbf{63.48}} & {\color[HTML]{3166FF} \textbf{66.84}} & 52.16  & {\color[HTML]{FE0000} \textbf{78.00}}   & {\color[HTML]{FE0000} \textbf{70.94}}  & {\color[HTML]{3166FF} \textbf{75.51}}  & {\color[HTML]{3166FF} \textbf{68.04}}  & {\color[HTML]{FE0000} \textbf{67.40}}  & {\color[HTML]{3166FF} \textbf{57.26}}  \\ \hline                

\end{tabular}
\caption{
ImageNet Results without AutoAugment~\cite{cubuk2019autoaugment}.
% Results on ImageNet without AutoAugment.
% DER w/o P means DER without pruning. 
% FOSTER B4 means the model before feature compression.
% FOSTER B4 means the model performance before feature compression.
}
% , Ours vs. state of the art. 
% DER w/o P means DER without pruning. FOSTER B4 means the model performance before feature compression.}
\label{tab:imagenet without da}
\end{table*}

\begin{table*}[t]
\centering
% \scalebox{1.0}
% {
\begin{tabular}{c|cccccc|cccc}
\hline
\multirow{3}{*}{Methods} & \multicolumn{6}{c|}{CIFAR100 B0}                                                        & \multicolumn{4}{c}{CIFAR100 B50}                        \\ \cline{2-11} 
                         & \multicolumn{2}{c}{5 steps} & \multicolumn{2}{c}{10 steps} & \multicolumn{2}{c|}{20 steps} & \multicolumn{2}{c}{5 steps} & \multicolumn{2}{c}{10 steps} \\ \cline{2-11} 
                         & Avg          & Last         & Avg          & Last          & Avg           & Last          & Avg          & Last         & Avg          & Last          \\ \hline
iCaRL ~\cite{icarl}     & 66.28                                 & 53.45                                 & 64.69                                 & 48.69                                 & 64.06                                 & 47.22     & 52.04  & 43.96  & 43.25  & 37.70                              \\
BiC ~\cite{wu2019large}       & 65.03                                 & 54.19                                 & 62.21                                 & 47.69                                 & 61.16                                 & 40.91                            & 58.36  & 44.40  & 55.71  & 42.93        \\
WA ~\cite{wa}        & 66.07                                 & 55.51                                 & 65.66                                 & 51.25                                 & 65.05                                 & 48.63                               & 60.86  & 51.99  & 55.84  & 48.12     \\
PODNet ~\cite{podnet}    & 63.69                                 & 50.18                                 & 55.98                                 & 38.04                                 & 48.68                                 & 29.10                             & 64.69  & 55.38  & {\color[HTML]{3166FF} \textbf{63.11}}  & 52.76       \\ 
DER w/o P ~\cite{der} & {\color[HTML]{FE0000} \textbf{71.35}}                                 & {\color[HTML]{FE0000} \textbf{63.55}}                                 & {\color[HTML]{FE0000} \textbf{71.02}}                                 & {\color[HTML]{FE0000} \textbf{60.18}}                                 & {\color[HTML]{FE0000} \textbf{69.75}}                                 & {\color[HTML]{FE0000} \textbf{57.22}} & {\color[HTML]{FE0000} \textbf{68.09}}  & {\color[HTML]{FE0000} \textbf{61.90}}  & {\color[HTML]{FE0000} \textbf{66.36}}  & {\color[HTML]{FE0000} \textbf{59.70}}   \\
FOSTER B4 ~\cite{foster} & 67.95 & 56.17 & 64.20 & 49.86 & 60.80 & 45.67                              & {\color[HTML]{3166FF} \textbf{65.47}}  & {\color[HTML]{3166FF} \textbf{57.27}}  & 61.81  & {\color[HTML]{3166FF} \textbf{53.17}}      \\
FOSTER ~\cite{foster}    & 65.77                                 & 54.94                                 & 62.70                                 & 49.14                                 & 59.94                                 & 45.04                             & 64.48  & 55.65  & 61.26  & 51.71       \\ \hline
MAFDRC      & {\color[HTML]{3166FF} \textbf{70.12}} & {\color[HTML]{3166FF} \textbf{59.05}} & {\color[HTML]{3166FF} \textbf{67.72}} & {\color[HTML]{3166FF} \textbf{54.24}} & {\color[HTML]{3166FF} \textbf{65.84}} & {\color[HTML]{3166FF} \textbf{50.76}} & 64.53  & 56.33  & 60.95  & 52.49   \\ \hline

\end{tabular}
\caption{
CIFAR100 Results without AutoAugment~\cite{cubuk2019autoaugment}.
% Results on CIFAR100 without AutoAugment.
}
% , Ours vs. state of the art. 
% DER w/o P means DER without pruning. FOSTER B4 means the model performance before feature compression.}
\label{tab:cifar-without da}
\end{table*}

To further mitigate the classification bias, the adjusted loss, logit adjustment (LA)~\cite{menon2020long}, is adopted by our method, as mentioned in Sec.~\ref{sec:ablation}.
Specifically,  an adjusting vector ${\gamma_t}$ is used to adjust the original classifier logit $\bar \ell_t$,
\begin{equation}
  \eta_t = \bar \ell_t + {\gamma _t},
  \label{eq:g}
\end{equation}
and the adjusted one, $\eta_t$, can be used to compute losses.
${\gamma_t}$ is computed as,
\begin{equation}
  \gamma _t^i = \log \left[ {{{(\frac{{{m_i}}}{{{m_1} + {m_2} +  \cdots  + {m_t}}})}^{tro}} } \right],
  \label{eq:r} %+ 1e - 12
\end{equation}
where ${m_i}$ is the sample number of the ${i^{th}}, i=\{1,...,t\}$ task in new data ${D}_t$ or memory buffer ${M}_t$.
${tro}$ is a hyperparameter of LA and is fixed at $1.2$.
% for almost experiments
% {\color{red}{
% except the ImageNet100 20 steps in 1.6.
% }}
As shown in Tab.~\ref{supp_tab:tro}, our method is not sensitive to different $tro$s.
% By adjusting the original classifier logit $\ell_t$ with ${\gamma_t}$,
% \begin{equation}
%   \eta_t = \ell_t + {\gamma _t},
%   \label{eq:g}
% \end{equation}
% the adjusted logit $\eta_t$ is used for classification. 

{\color{red}{

% \noindent {\bf The vision transformer (VIT) based DRC} \quad
\noindent {\bf The Vision Transformer (VIT) Backbone} \quad
The DyTox ViT encoder replaces ResNet as the feature extractor, resulting in MAF(ViT) and MAFDRC(ViT). As shown in ~\ref{tab:transformer}, {DRC clearly boosts the baseline results and performs similarly to DyTox.} 

% \noindent {\bf Results under different class orders} \quad
\noindent {\bf Averaged Results of Three Runs} \quad
% The averaged results of three runs are shown in ~\ref{tab:imagenet_three_orders} and ~\ref{tab:cifar_three_orders}.
As shown in ~\ref{tab:imagenet_three_orders} and ~\ref{tab:cifar_three_orders},
% They are consistent with the results in Tab.~1 and Tab.~2.
the proposed method still achieves the SOTA-level performance and is statistically better than its counterparts.
% Similar to the main results, the proposed method consistently achieve SOTA-level performance and is statistically better than its counterparts.
% We will replace it using three runs in the main text.
% More results will be added.
% Will add these results.

}}

\begin{table}[t]
\centering
\footnotesize
% \scalebox{0.7}{
\begin{tabular}{c|cccc}
\hline
\multirow{3}{*}{Methods} & \multicolumn{4}{c}{CIFAR100 B0}                                                              \\ \cline{2-5} 
                         & \multicolumn{2}{c}{5 steps} & \multicolumn{2}{c}{10 steps}  \\ \cline{2-5} 
                         & Avg          & Last         & Avg           & Last                  \\ \hline

MAF(ViT)                & 63.83        & 44.84        & 58.58         & 36.69               \\
DyTox$^{\dag}$               & 71.78        & 61.31        & 69.63         & 55.65                \\
MAFDRC(ViT)                  & 72.48        & 62.78        & 69.49         & 56.13             \\ 
% MAFDRC-resnet                   & 74.87       & 73.97        & 71.75        & 71.15           \\ 
\hline
\end{tabular}
% \vspace{-0.1cm}
\caption{CIL results with ViT-based feature extractor. $^{\dag}$ indicates the DyTox results are reproduced under the same setting (e.g., data augmentation and trained on a single GPU) with others.
\textcolor{red}{20-step results?}
}
% \caption{The results of different model.}
\label{tab:transformer}
\end{table}

\begin{table}[t]
\centering
\scalebox{0.6}
{
\begin{tabular}{c|cccccc}
\hline
                          & \multicolumn{2}{c}{5 steps}                                                    & \multicolumn{2}{c}{10 steps}                                                  & \multicolumn{2}{c}{20 steps}                                                   \\ \cline{2-7} 
\multirow{-2}{*}{Methods} & Avg                                   & Last                                  & Avg                                   & Last                                  & Avg                                   & Last                                  \\ \hline

DER w/o P                       & {\color[HTML]{3166FF} \textbf{${81.64}_{\pm 0.62}$}} & {\color[HTML]{3166FF} \textbf{${75.23}_{\pm 0.69}$}} & {\color[HTML]{3166FF} \textbf{${79.54}_{\pm 1.10}$}} & {\color[HTML]{3166FF} \textbf{${70.55}_{\pm 0.52}$}} & {\color[HTML]{FE0000} \textbf{${78.12}_{\pm 0.11}$}} & {\color[HTML]{FE0000} \textbf{${70.91}_{\pm 0.60}$}} \\
FOSTER B4                 & ${80.31}_{\pm 0.67}$                                 & ${73.28}_{\pm 0.61}$                                 & ${77.10}_{\pm 0.51}$                                 & ${67.93}_{\pm 0.87}$                                 & ${74.39}_{\pm 0.21}$                                 & ${62.65}_{\pm 0.66}$                                 \\
FOSTER                    & ${79.35}_{\pm 0.94}$                                 & ${71.87}_{\pm 0.48}$                                 & ${76.51}_{\pm 0.28}$                                 & ${67.09}_{\pm 0.53}$                                 & ${74.20}_{\pm 0.30}$                                 & ${62.78}_{\pm 0.37}$                                 \\
MAFDRC                    & {\color[HTML]{FE0000} \textbf{${82.31}_{\pm 0.33}$}} & {\color[HTML]{FE0000} \textbf{${76.05}_{\pm 0.30}$}} & {\color[HTML]{FE0000} \textbf{${79.78}_{\pm 0.28}$}} & {\color[HTML]{FE0000} \textbf{${70.56}_{\pm 0.39}$}} & {\color[HTML]{3166FF} \textbf{${75.63}_{\pm 0.25}$}} & {\color[HTML]{3166FF} \textbf{${64.14}_{\pm 0.71}$}} \\ \hline

\end{tabular}}
\caption{
% Three runs results (avg acc. ± std err.) on ImageNet100.
Three runs results (averaged accuracy ± standard error) on ImageNet100 B0.
Red indicates the best performance and blue indicates the second best results.
% Results on ImageNet100 on 3 runs.
}
% , Ours vs. state of the art. 
% DER w/o P means DER without pruning. FOSTER B4 means the model performance before feature compression.}
\label{tab:imagenet_three_orders}
\end{table}

\begin{table}[t]
\centering
\scalebox{0.6}
% \small
{
\begin{tabular}{c|cccccc}
\hline
                          & \multicolumn{2}{c}{5 steps}                                                    & \multicolumn{2}{c}{10 steps}                                                  & \multicolumn{2}{c}{20 steps}                                                   \\ \cline{2-7} 
\multirow{-2}{*}{Methods} & Avg                                   & Last                                  & Avg                                   & Last                                  & Avg                                   & Last                                  \\ \hline

DER w/o P                       & {\color[HTML]{FE0000} \textbf{${75.19}_{\pm 0.61}$  }} & {\color[HTML]{FE0000} \textbf{${68.78}_{\pm 0.34}$  }} & {\color[HTML]{FE0000} \textbf{${74.84}_{\pm 0.81}$}} & {\color[HTML]{FE0000} \textbf{${65.74}_{\pm 0.55}$}} & {\color[HTML]{FE0000} \textbf{${73.85}_{\pm 0.64}$}} & {\color[HTML]{FE0000} \textbf{${62.92}_{\pm 0.73}$}} \\
FOSTER B4                 & ${73.96}_{\pm 0.77}$                                & ${64.78}_{\pm 0.71}$                                 & ${72.91}_{\pm 0.74}$                                 & ${61.54}_{\pm 0.90}$                                 & ${70.40}_{\pm 0.57}$                                 & ${56.87}_{\pm 0.66}$                                 \\
FOSTER                    & ${71.85}_{\pm 1.01}$                                 & ${62.58}_{\pm 0.99}$                                 & ${71.58}_{\pm 0.62}$                                 & ${60.39}_{\pm 0.28}$                                 & ${69.41}_{\pm 0.55}$                                 & ${56.09}_{\pm 0.93}$                                \\
MAFDRC                    & {\color[HTML]{3166FF} \textbf{${74.81}_{\pm 0.05}$}} & {\color[HTML]{3166FF} \textbf{${66.16}_{\pm 0.26}$}} & {\color[HTML]{3166FF} \textbf{${73.85}_{\pm 0.16}$}} & {\color[HTML]{3166FF} \textbf{${61.98}_{\pm 0.25}$}} & {\color[HTML]{3166FF} \textbf{${71.93}_{\pm 0.34}$}} & {\color[HTML]{3166FF} \textbf{${57.43}_{\pm 0.34}$}} \\ \hline

\end{tabular}}
\caption{Three runs results (averaged accuracy ± standard error) on CIFAR100 B0. Red indicates the best performance and blue indicates the second best results.}
% , Ours vs. state of the art. 
% DER w/o P means DER without pruning. FOSTER B4 means the model performance before feature compression.}
\label{tab:cifar_three_orders}
\end{table}